\documentclass[10pt,twocolumn,letterpaper]{article}

\usepackage[pagenumbers]{cvpr} 

\usepackage{graphicx}
\usepackage{xcolor}         
\usepackage[acronym]{glossaries}

\usepackage{algorithm}
\usepackage{algpseudocode}
\usepackage{enumitem}
\usepackage{amssymb}

\usepackage{xcolor}
\usepackage{graphicx}
\usepackage{bm}
\newcommand{\vect}[1]{\boldsymbol{\mathbf{#1}}}
\usepackage{caption}
\usepackage{amsmath,amssymb}
\usepackage{tabularray}
\UseTblrLibrary{booktabs}
\usepackage{makecell}
\usepackage{multirow}
\usepackage{subcaption}    
\usepackage[table]{xcolor}
\definecolor{darkgreen}{RGB}{0,150,0}
\definecolor{darkred}{RGB}{150,10,10}
\usepackage{stackengine}

\definecolor{cvprblue}{rgb}{0.21,0.49,0.74}
\usepackage[pagebackref,breaklinks,colorlinks,allcolors=cvprblue]{hyperref}

\def\paperID{20315} 
\def\confName{CVPR}
\def\confYear{2026}

\title{The Invisible Gorilla Effect in Out-of-distribution Detection}

\author{Harry Anthony$^{\textrm{1}}$, 
        Ziyun Liang$^{\textrm{1,*}}$,
        Hermione Warr$^{\textrm{1,*}}$,
        Konstantinos Kamnitsas$^{\textrm{1}}$ \\
Department of Engineering Science, University of Oxford, Oxford, UK \\
{\tt\small harry.anthony@eng.ox.ac.uk} \\
{\small * Equal second-author contribution}} 

\begin{document}

\maketitle
\begin{abstract}
Deep Neural Networks achieve high performance in vision tasks by learning features from regions of interest (ROI) within images, but their performance degrades when deployed on out-of-distribution (OOD) data that differs from training data. This challenge has led to OOD detection methods that aim to identify and reject unreliable predictions. Although prior work shows that OOD detection performance varies by artefact type, the underlying causes remain underexplored. To this end, we identify a previously unreported bias in OOD detection: for hard-to-detect artefacts (near-OOD), detection performance typically improves when the artefact shares visual similarity (e.g. colour) with the model’s ROI and drops when it does not - a phenomenon we term the Invisible Gorilla Effect. For example, in a skin lesion classifier with red lesion ROI, we show the method Mahalanobis Score achieves a 31.5\% higher AUROC when detecting OOD red ink (similar to ROI) compared to black ink (dissimilar) annotations. We annotated artefacts by colour in 11,355 images from three public datasets (e.g. ISIC) and generated colour-swapped counterfactuals to rule out dataset bias. We then evaluated 40 OOD methods across 7 benchmarks and found significant performance drops for most methods when artefacts differed from the ROI. Our findings highlight an overlooked failure mode in OOD detection and provide guidance for more robust detectors. Code and annotations are available at: \url{https://github.com/HarryAnthony/Invisible\_Gorilla\_Effect}.
\end{abstract}    
\hypertarget{p1}{}

\section{Introduction}
\label{sec:intro}

\noindent Deep neural networks (DNNs) have achieved expert-level accuracy in computer vision (CV) tasks across a range of applications, including high-risk settings such as medical imaging \cite{esteva_dermatologist-level_2017,gulshan_performance_2019} and autonomous driving \cite{janai2020computer}. Models achieve this strong performance by learning discriminative features localised in regions of interest (ROI) within images. However, DNNs typically lose their expert-level performance when applied to data that differs significantly from the training data \cite{perone2019unsupervised,amodei_concrete_2016}, known as out-of-distribution (OOD) data. In real-world settings, models are likely to encounter OOD inputs due to the inherent variability of real-world data and the challenges of assembling comprehensive training datasets - especially in privacy-sensitive domains like medical imaging. This challenge has led to the development of out-of-distribution detection methods, which aim to detect and filter out unreliable predictions on OOD inputs - facilitating safer application of neural networks for high-risk applications. Recent US and EU regulatory guidance have highlighted the need for machine learning systems to handle OOD inputs \cite{fda2025guidance,eu2021aiact,gonzalez2024regulating}, underscoring the practical relevance of this problem.

This has motivated the development of numerous OOD detection methods. For a DNN trained on a primary task (e.g. image classification) - hereafter referred to as the primary model - OOD detection methods are typically divided into two categories \cite{gao2024comprehensive}: \emph{internal methods}, which use the primary model’s outputs or parameters, and \emph{external methods}, which are external to the primary model. External methods employ a variety of models, including reconstruction-based (e.g. DDPM \cite{graham2023denoising}), density-based (e.g. flow-based models \cite{dinh2017density}) and classification approaches (e.g. Deep SVDD \cite{ruff2018deep}). Internal methods are commonly divided into two groups \cite{molnar2020interpretable}: \emph{ad-hoc} methods, approaches that require altering the primary model architecture or training to incorporate OOD-awareness into the learning process (e.g. Bayesian neural networks \cite{blundell2015weight} and Rotation Prediction \cite{hendrycks2019using}), and \emph{post-hoc} methods, which can be applied to pre-trained models without modification. Post-hoc methods are commonly broken down into two categories: \emph{classification-based methods}, which use the model’s output or penultimate layer for OOD detection (e.g. MCP \cite{hendrycks_baseline_2016}), and \emph{feature-based methods}, which utilise hidden-layer representations (e.g. Mahalanobis Score \cite{lee2018simple}).

\begin{figure*}[t] 
    \centering
    \stackinset{l}{351pt}{b}{108pt}%
  {\rotatebox{0}{\hyperref[sec:bib]{\phantom{\rule{9pt}{6pt}}}}}{}%
   \stackinset{l}{345pt}{b}{125pt}%
  {\rotatebox{5}{\hyperref[sec:bib]{\phantom{\rule{9pt}{6pt}}}}}{}%
     \stackinset{l}{349pt}{b}{143pt}%
  {\rotatebox{10}{\hyperref[sec:bib]{\phantom{\rule{9pt}{6pt}}}}}{}%
       \stackinset{l}{319pt}{b}{145pt}%
  {\rotatebox{20}{\hyperref[sec:bib]{\phantom{\rule{9pt}{6pt}}}}}{}%
        \stackinset{l}{335pt}{b}{177pt}%
  {\rotatebox{20}{\hyperref[sec:bib]{\phantom{\rule{6pt}{6pt}}}}}{}%
          \stackinset{l}{305pt}{b}{168pt}%
  {\rotatebox{20}{\hyperref[sec:bib]{\phantom{\rule{6pt}{6pt}}}}}{}%
        \stackinset{l}{290pt}{b}{173pt}%
  {\rotatebox{45}{\hyperref[sec:bib]{\phantom{\rule{9pt}{6pt}}}}}{}%
         \stackinset{l}{283pt}{b}{189pt}%
  {\rotatebox{45}{\hyperref[sec:bib]{\phantom{\rule{9pt}{6pt}}}}}{}%
           \stackinset{l}{263pt}{b}{188pt}%
  {\rotatebox{75}{\hyperref[sec:bib]{\phantom{\rule{9pt}{6pt}}}}}{}%
             \stackinset{l}{250pt}{b}{179pt}%
  {\rotatebox{90}{\hyperref[sec:bib]{\phantom{\rule{9pt}{6pt}}}}}{}%
            \stackinset{l}{237pt}{b}{188pt}%
  {\rotatebox{100}{\hyperref[sec:bib]{\phantom{\rule{9pt}{6pt}}}}}{}%
            \stackinset{l}{222pt}{b}{180pt}%
  {\rotatebox{100}{\hyperref[sec:bib]{\phantom{\rule{9pt}{6pt}}}}}{}%
        \stackinset{l}{210pt}{b}{171pt}%
  {\rotatebox{110}{\hyperref[sec:bib]{\phantom{\rule{9pt}{6pt}}}}}{}%
          \stackinset{l}{197pt}{b}{166pt}%
  {\rotatebox{110}{\hyperref[sec:bib]{\phantom{\rule{9pt}{6pt}}}}}{}%
            \stackinset{l}{186pt}{b}{160pt}%
  {\rotatebox{130}{\hyperref[sec:bib]{\phantom{\rule{9pt}{6pt}}}}}{}%
            \stackinset{l}{177pt}{b}{147pt}%
  {\rotatebox{130}{\hyperref[sec:bib]{\phantom{\rule{9pt}{6pt}}}}}{}%
            \stackinset{l}{167pt}{b}{137pt}%
  {\rotatebox{150}{\hyperref[sec:bib]{\phantom{\rule{9pt}{6pt}}}}}{}%
              \stackinset{l}{145pt}{b}{135pt}%
  {\rotatebox{150}{\hyperref[sec:bib]{\phantom{\rule{9pt}{6pt}}}}}{}%
            \stackinset{l}{140pt}{b}{118pt}%
  {\rotatebox{160}{\hyperref[sec:bib]{\phantom{\rule{9pt}{6pt}}}}}{}%
              \stackinset{l}{160pt}{b}{100pt}%
  {\rotatebox{175}{\hyperref[sec:bib]{\phantom{\rule{6pt}{6pt}}}}}{}%
                \stackinset{l}{161pt}{b}{86pt}%
  {\rotatebox{185}{\hyperref[sec:bib]{\phantom{\rule{6pt}{6pt}}}}}{}%
            \stackinset{l}{142pt}{b}{69pt}%
  {\rotatebox{195}{\hyperref[sec:bib]{\phantom{\rule{6pt}{6pt}}}}}{}%
              \stackinset{l}{167pt}{b}{57pt}%
  {\rotatebox{225}{\hyperref[sec:bib]{\phantom{\rule{9pt}{6pt}}}}}{}%
              \stackinset{l}{172pt}{b}{45pt}%
  {\rotatebox{245}{\hyperref[sec:bib]{\phantom{\rule{9pt}{6pt}}}}}{}%
            \stackinset{l}{165pt}{b}{21pt}%
  {\rotatebox{245}{\hyperref[sec:bib]{\phantom{\rule{9pt}{6pt}}}}}{}%
              \stackinset{l}{187pt}{b}{21pt}%
  {\rotatebox{245}{\hyperref[sec:bib]{\phantom{\rule{9pt}{6pt}}}}}{}%
                \stackinset{l}{200pt}{b}{12pt}%
  {\rotatebox{255}{\hyperref[sec:bib]{\phantom{\rule{9pt}{6pt}}}}}{}%
                  \stackinset{l}{210pt}{b}{3pt}%
  {\rotatebox{255}{\hyperref[sec:bib]{\phantom{\rule{9pt}{6pt}}}}}{}%
            \stackinset{l}{224pt}{b}{7pt}%
  {\rotatebox{255}{\hyperref[sec:bib]{\phantom{\rule{9pt}{6pt}}}}}{}%
              \stackinset{l}{240pt}{b}{1pt}%
  {\rotatebox{255}{\hyperref[sec:bib]{\phantom{\rule{9pt}{6pt}}}}}{}%
                \stackinset{l}{254pt}{b}{7pt}%
  {\rotatebox{285}{\hyperref[sec:bib]{\phantom{\rule{9pt}{6pt}}}}}{}%
            \stackinset{l}{273pt}{b}{1pt}%
  {\rotatebox{295}{\hyperref[sec:bib]{\phantom{\rule{9pt}{6pt}}}}}{}%
              \stackinset{l}{283pt}{b}{11pt}%
  {\rotatebox{295}{\hyperref[sec:bib]{\phantom{\rule{9pt}{6pt}}}}}{}%
             \stackinset{l}{290pt}{b}{23pt}%
  {\rotatebox{315}{\hyperref[sec:bib]{\phantom{\rule{9pt}{6pt}}}}}{}%
            \stackinset{l}{300pt}{b}{31pt}%
  {\rotatebox{335}{\hyperref[sec:bib]{\phantom{\rule{9pt}{6pt}}}}}{}%
              \stackinset{l}{332pt}{b}{16pt}%
  {\rotatebox{335}{\hyperref[sec:bib]{\phantom{\rule{9pt}{6pt}}}}}{}%
                \stackinset{l}{335pt}{b}{40pt}%
  {\rotatebox{345}{\hyperref[sec:bib]{\phantom{\rule{9pt}{6pt}}}}}{}%
                \stackinset{l}{342pt}{b}{58pt}%
  {\rotatebox{345}{\hyperref[sec:bib]{\phantom{\rule{9pt}{6pt}}}}}{}%
                  \stackinset{l}{351pt}{b}{73pt}%
  {\rotatebox{350}{\hyperref[sec:bib]{\phantom{\rule{9pt}{6pt}}}}}{}%
                    \stackinset{l}{346pt}{b}{90pt}%
  {\rotatebox{355}{\hyperref[sec:bib]{\phantom{\rule{9pt}{6pt}}}}}{}%
    {\includegraphics[width=\textwidth]{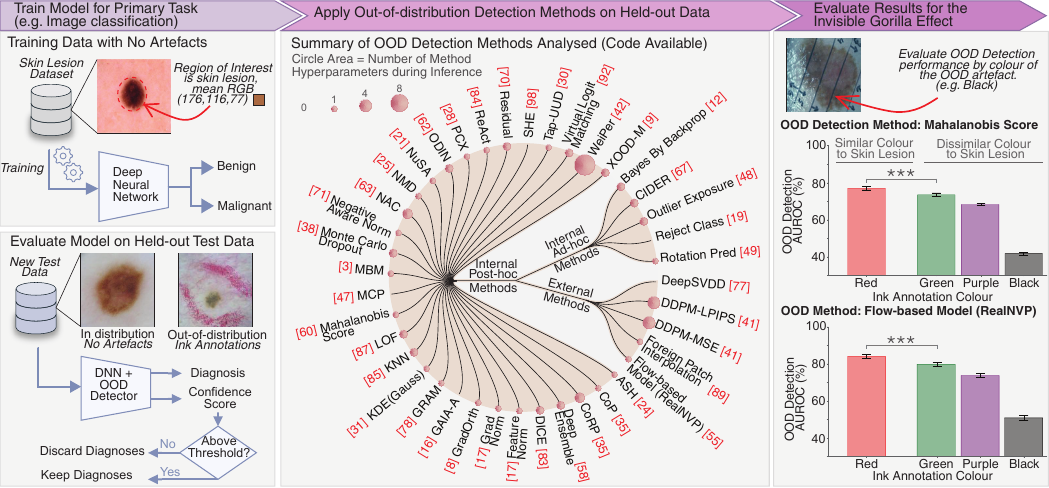}} 
    \caption{\textbf{Invisible Gorilla Effect.} (Left) A DNN (e.g. a ResNet-18 skin lesion classifier) is trained for a primary task, where the region of interest (ROI) is the skin lesion, with a mean RGB of $(176,116,77)$ across the dataset. Once trained, the model is deployed on held-out data, including OOD data containing an unseen ink artefact. (Centre) Visualisation of the OOD detection methods analysed in this study, categorised as Internal Post-hoc, Internal Ad-hoc and External. Circle area reflects the number of method hyperparameters evaluated. (Right) We evaluated OOD detection AUROC on artefacts with colours similar (red) and dissimilar (green, purple, black) to the model’s ROI. A statistically significant AUROC drop ($p<10^{-5}$, Wilcoxon signed-rank) is observed for Mahalanobis Score and RealNVP methods on dissimilar-colour artefacts - an instance of the Invisible Gorilla Effect. Error bars denote 95\% confidence intervals over 25 random seeds. } 
    \label{fig:IGE}
\end{figure*}

Although many OOD detection methods have been proposed, prior work has shown that strong performance on one type of OOD artefact can often fail to generalise to other artefacts \cite{tajwar2021no,anthony_use_2023,gutbrod2025openmibood} - yet the underlying causes remain largely unexplored. A key challenge in real-world OOD detection is that the types of inputs a model may encounter are not known \emph{a priori}, highlighting the need for methods that generalise across a variety of distribution shifts. Therefore, the community would benefit from a deeper analysis of why such performance drops occur, and how to design OOD detection systems that are robust across diverse tasks. To this end, we identify a previously unreported bias in OOD detection for hard-to-detect artefacts (known as near-OOD). We observe that the OOD detection performance typically depends on visual similarity (e.g. colour and brightness) of an OOD artefact to the model’s ROI, improving when similarity is high to the ROI and degrading when it is not, a phenomenon we term the \emph{Invisible Gorilla Effect}. For example, in a dermatology classifier with red lesion ROI, we find that images containing red OOD artefacts (e.g. ink, colour charts) yield a significant improvement in detection performance across most OOD methods, compared to the same artefacts in visually dissimilar colours  (e.g. green, black).

The name of this effect references the famous \emph{Invisible Gorilla Experiment} \cite{chabris2011invisible}, in which participants asked to count basketball passes between players in white shirts - while ignoring passes between players in black shirts - often failed to notice a person in a black gorilla suit walking through the scene. The experiment was used to illustrate the cognitive effect called \emph{Inattentive Blindness}, where individuals focusing on a primary task can ignore unexpected stimuli due to a lack of attention. Subsequent research has shown that unexpected objects are more likely to be perceived when they closely resemble the attended target \cite{carpenter2001sights,most2001not}, thus reducing the chance of inattentive blindness. Our findings suggest that an analogous effect exists in OOD detection, where artefacts visually similar to the model’s ROI are significantly more likely to be detected.


This study focuses on colour similarity as a controlled, practically relevant factor, as colour artefacts are common and can be varied independently of shape or texture. We make the following contributions:
\begin{itemize}
    \item We identify the \textit{Invisible Gorilla Effect} in OOD detection, completing an extensive study of 40 OOD detection methods (3795 hyperparameter settings) across 7 benchmarks and 3 network architectures.
    \item We annotated 11,355 images from public datasets by artefact colour, and generated colour-swapped counterfactuals of OOD artefacts to rule out dataset bias. Annotations are public on project's GitHub repository.
    
    \item We observed that feature-based OOD methods suffer larger drops in detection performance than confidence-based methods as artefact colour varies, and we used PCA-based latent analysis to show that these colour shifts align with high-variance directions in the latent space.
    
    \item We identified a \textit{nuisance subspace} that captures high-variance, colour-sensitive directions in the latent space, and showed that projecting features orthogonally to this subspace reduces the Invisible Gorilla Effect for several feature-based methods.

\end{itemize}
\section{Background}
\label{sec:background}

\begin{table*}[ht!]
\centering
\footnotesize            
\begin{tabular}{lllllll}
\toprule
\textbf{Dataset} & \textbf{Training Data Desc.} & \textbf{\# Training Images} & \textbf{\# ID Test Images} & \textbf{OOD Data Desc.} & \textbf{\# OOD Images} & \textbf{\# OOD Type} \\
\midrule
CheXpert & No Support Devices & 5\,213 & 1\,282 & Synthetic squares & 1\,282 & Synthetic Covariate \\ 
\midrule
\multirow{2}{*}{ISIC} 
  & \multirow{2}{*}{\makecell[l]{        Planar images \\[-2pt]%
        with no rulers,\\[-2pt]%
        ink or colour charts }} 
  & \multirow{2}{*}{25\,266} 
  & \multirow{2}{*}{6\,316} 
  & Colour charts  & 8\,964 & Real Covariate \\ \cmidrule{5-7} 
 &   &  &  & Ink annotations & 2\,358 & Real Covariate \\
 \midrule
MVTec & Metal nuts with no ink & 286 & 71 & Ink artefacts & 16 & Semantic (new defect) \\ 
\midrule
MVTec & Pills with no ink & 233 & 58 & Ink artefacts & 18 & Semantic (new defect) \\ 

\bottomrule
\end{tabular}
\caption{Summary of Training data, ID test data and OOD test data used for OOD detection evaluation.}
\label{tbl:datasets}
\end{table*}

\subsection{Primer on Out-of-distribution Detection}
Consider a model's input, $\mathbf{x} \in \mathcal{X}$ and a label $\vect{y} \in \mathcal{Y}$, related by  $p(\mathbf{x},\vect{y})$. A neural network, $f$, is trained on a dataset $\mathcal{D}_{\text {train}} = \{ (\vect{x_n}, \vect{y_n}) \}_{n=1}^N \subset \mathcal{X} \times \mathcal{Y}$ to map from an input set $\mathcal{X}$ to a label set $\mathcal{Y}$. Once the model is trained, it can be deployed on unseen data. A distribution shift occurs when $p_{\text{train}}(\mathbf{x},\vect{y}) \neq p_{\text{test}}(\mathbf{x},\vect{y})$ \cite{valiant_theory_1984}. In OOD detection, the two common types are: covariate shift, where the input distribution changes but label semantics do not ($p_{\text{train}}(\mathbf{x}) \neq p_{\text{test}}(\mathbf{x})$ while $p_{\text{train}}(y | \mathbf{x}) = p_{\text{test}}(y | \mathbf{x})$), and semantic shift, where novel classes appear at test time ($p_{\text{train}}(y | \mathbf{x}) \neq p_{\text{test}}(y | \mathbf{x})$ while $p_{\text{train}}(\mathbf{x}) \approx p_{\text{test}}(\mathbf{x}))$  \cite{valiant_theory_1984}.



For this study, we operationalised our definition of OOD under three regimes (real semantic shifts, real covariate shifts and synthetic covariate shifts) for medical and industrial datasets. We emphasise that OODness is defined here operationally rather than with a fixed quantitative boundary, since any absolute boundary cannot be defined \emph{a priori} as this would presuppose a perfect OOD detector.
\begin{itemize}
    \item \textbf{Semantic OOD:} Inputs from classes unseen during training were defined as OOD, $\mathcal{Y}_{\text{train}} \cap \mathcal{Y}_{\text{test}} = \emptyset$. 
    \item \textbf{Real covariate OOD:} Inputs drawn from the same acquisition domain but containing naturally occurring visual artefacts (e.g. ink annotations) were defined as OOD. Unlike conventional domain shifts (e.g. data from a new device or site), these artefacts are not identifiable from medical metadata and therefore require image-based OOD detection. This setting captures naturally occurring, metadata-invisible variations that degrade model performance (see Sec \ref{Sec:IGE}) in real clinical workflows.
   \item \textbf{Synthetic covariate OOD:} Controlled perturbations were applied to ID samples to create synthetic OOD inputs. These were not meant to reflect realistic deployment conditions, but were used as diagnostic tools to systematically vary visual similarity and probe how the model’s ROI affects OOD detection.
\end{itemize}

Each OOD detection method produces a scoring function, $\mathcal{S}(\mathbf{x})$, for each input, where higher value indicate greater confidence of belonging to the in-distribution (ID). The performance of the scoring function for OOD detection can be evaluated against the ground-truth ID/OOD splits, using metrics such as AUROC, AUCPR and FPR@95 \cite{hendrycks_baseline_2016}.

OOD detection is often framed on a spectrum from \textit{near-OOD} to \textit{far-OOD} \cite{scheirer2012toward}. Although studies differ significantly in how they operationalise these terms, near-OOD typically denotes difficult OOD detection tasks. Underlying this framing is the assumption of a monotonic relationship between similarity and detectability: as OOD data become more similar to the training distribution, detection performance should decline \cite{ren2021simple,rafiee2022self,du2024does,cheng2025average,zadorozhny2022out,liu2024neuron,fang2022out,yuan2023devil}. However, our findings challenge this assumption by revealing a bias case where OOD samples that are more visually similar to the model’s region of interest are, in fact, easier to detect than less similar ones. This \emph{counterintuitive} effect suggests that OOD task difficulty is not a simple function of global similarity to the training data, but depends on factors such as what model attends to and how that interacts with the detection method.


\subsection{Related Works}
While OOD detection methods have shown large performance variation across datasets and artefact types in the literature, the underlying causes of these variations remain largely unexplored. A recent study used counterfactuals (inputs with and without OOD artefacts) to show that real-world OOD artefacts can lead to high-confidence predictions (large logits) comparable to ID inputs, reducing the detection performance of confidence-based methods \cite{anthony_evaluating_2025}. This introduces variability in detection performance, as artefacts that correlate with a particular class and lead to high-confidence predictions will be harder to detect. Other studies have shown that a key failure case for density- and generative-based methods (where OOD inputs receive higher likelihoods than ID inputs) can arise when background regions dominate the low-level statistics of an image \cite{ren2019likelihood,caterini2022entropic,kamkari2024geometric}. Other studies have investigated the factors influencing performance variation in feature-based methods (e.g. Mahalanobis score), showing that the choice of feature layer significantly affects the detection performance for a given artefact \cite{lambert2023multi}. It has been shown that the optimal layer for detecting one type of OOD artefact with Mahalanobis Score can be the least effective for detecting another \cite{anthony_use_2023}. 


\section{Methods and Materials}
\label{sec:methods}

\subsection{Datasets and Implementation}
We conducted our analysis on three public datasets: CheXpert \cite{irvin_chexpert_2019} (chest X-rays), ISIC \cite{codella_skin_2019} (dermatology) and MVTec-AD \cite{bergmann2019mvtec} (industrial inspection). A summary of the datasets is given in Table~\ref{tbl:datasets}. Models were trained using 25 random seeds  across five 5-fold cross-validation splits, using ResNet18 \cite{he_deep_2016}, VGG16 \cite{simonyan_very_2015} and ViT-B/32 \cite{dosovitskiy_image_2021}. 

\begin{itemize}
    \item \textbf{CheXpert:} Models were trained on frontal chest X-rays without support devices (artefacts removed to prevent potential biases) such as pacemakers (annotations from \cite{anthony_use_2023}) to perform binary classification between cardiomegaly and no findings. 
    \item \textbf{ISIC:} We manually annotated the dataset to identify planar images with and without non-diagnostic artefacts, such as colour charts, rulers and ink annotations (annotations to be released with this work). Models were then trained on artefact-free images to classify skin lesions as either malignant or benign. 
    \item \textbf{MVTec-AD:} We used two benchmarks: one with four metal nut classes (no defect, bent, flipped, scratch) and another with five pill classes (no defect, contamination, crack, faulty imprint, scratch).
\end{itemize}
 The per-channel mean and standard deviation were computed over the entire training dataset and used to normalise all images. Computational hardware for experiments is detailed in the supplementary material.

\begin{figure*} 
    \centering
    \includegraphics[width=\linewidth]{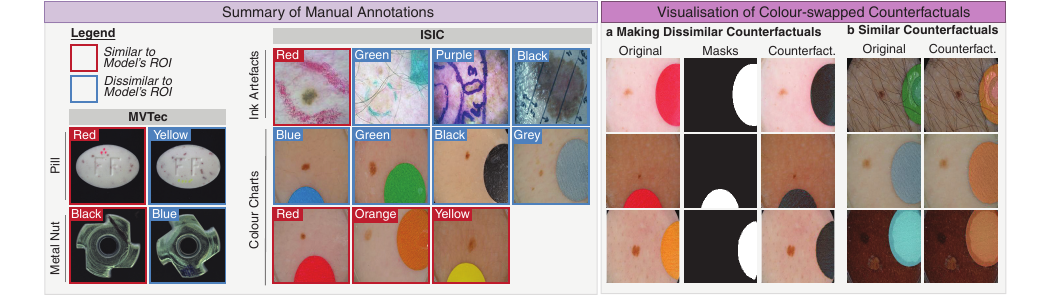} 
    \caption{(Left) Visualisation of annotated images from MVTec and ISIC, where red borders indicate artefacts visually similar to the model’s region of interest (ROI), and blue borders indicate dissimilar artefacts. (Right) Examples of colour-swapped counterfactuals for colour charts for both dissimilar and similar artefacts, showing the original image, artefact mask and the resulting counterfactual image. }
    \label{fig:annotation_summary}
\end{figure*}

The held-out portion of the training data was used as the ID test set. For CheXpert, synthetic squares covering 1\% of the image area were added at random positions to create synthetic covariate OOD. For ISIC, images with colour charts or ink annotations were used as OOD data, with artefacts manually annotated by colour (Figure~\ref{fig:annotation_summary}). We restricted our analysis to charts covering $<$10\% of the image, as larger artefacts yield uniformly high AUROC across all colours. For MVTec, the OOD images contained unseen ink artefacts, which were annotated by colour. Images with multiple artefact colours were excluded to isolate the effect of individual colours. Artefact counts by colour are provided in the supplementary material, and all annotations are available on the project's GitHub repository.

To analyse the Invisible Gorilla Effect, we classified OOD artefacts as similar or dissimilar based on linear Euclidean RGB distance to the model’s ROI. The model’s ROI and OOD artefacts were segmented using Segment Anything Model (SAM) \cite{kirillov_segment_2023} with interactive prompts, or using predefined ground-truth masks (only for ink annotations in MVTec \cite{bergmann2019mvtec}). Mean colours were then computed for each segment, with details in the supplementary. The similar/dissimilar threshold was benchmark-specific. 


To rule out dataset bias as the cause of the Invisible Gorilla Effect, we also generated synthetic colour-swapped counterfactuals of the colour charts (see Figure~\ref{fig:annotation_summary}). Using the segmentation masks described earlier, we computed the mean RGB values of each colour chart and then recoloured it with per-channel mean shifting, preserving pixel-level variance and texture. We changed visually similar charts (red, orange, yellow) to black, generating dissimilar counterfactuals as shown in Figure \ref{fig:annotation_summary}a. Black was chosen because it produced the lowest OOD detection performance on real data, with a mean RGB value of (66,61,60). Dissimilar charts (green, blue, black, grey) were changed to the average skin lesion colour (176,116,77), as shown in Figure \ref{fig:annotation_summary}b. We then evaluated OOD detection performance on these modified images.


Finally, we assessed how the model's ROI influences the Invisible Gorilla Effect. In the CheXpert benchmark, models were trained to classify cardiomegaly - an enlarged heart that appears as a hyperintense region in chest X-rays. Using CheXmask segmentation masks \cite{gaggion2024chexmask}, we created synthetic counterfactuals by scaling heart-region pixel intensities by one-third, producing a hypointense appearance while preserving texture (visualised in Figure~\ref{fig:Chex}). To smooth mask boundaries, we applied a Gaussian filter (kernel size 7, std 0.75). After dataset-wide per-channel mean and standard deviation were recomputed, models were trained on the counterfactual dataset and the OOD detection was evaluated with synthetic OOD artefacts.

\subsection{Out-of-distribution Detection Methods}

To enable a systematic evaluation of the Invisible Gorilla Effect, we conducted a broad empirical study across 40 OOD detection methods, visualised in Figure \ref{fig:IGE}. For each method, we explored an extensive set of hyperparameter configurations, such as varying the model layer at which feature-based methods are applied, totalling 3795 configurations (detailed in the supplementary material).

We evaluated 30 internal post-hoc methods, grouped into confidence-based and feature-based approaches as defined in the Introduction. In this work, we define confidence-based methods as those whose OOD scoring function utilises the model’s output or penultimate layer - even if they also use earlier features, as in ViM \cite{wang2022vim}. We evaluated 13 confidence-based methods: ASH \cite{djurisicextremely}, Deep Ensemble \cite{lakshminarayanan_simple_2017}, DICE \cite{sun2022dice}, GAIA-A \cite{chen2023gaia} , GradNorm \cite{chen2018gradnorm}, GradOrth \cite{behpour2023gradorth}, MCP \cite{hendrycks_baseline_2016}, MC-Dropout \cite{gal2016dropout}, ODIN \cite{liang2018enhancing}, ReAct \cite{sun2021react}, SHE \cite{zhang2022out}, ViM \cite{wang2022vim} and WeiPer \cite{granz2024weiper}. In addition, we studied 17 feature-based methods: CoP \cite{fang2024kernel}, CoRP \cite{fang2024kernel}, FeatureNorm \cite{dhamija2018reducing}, GRAM \cite{sastry2020detecting}, KDE (Gaussian) \cite{erdil2021task}, Deep KNN \cite{sun2022out}, LOF \cite{szyc2021out}, Mahalanobis Score \cite{lee2018simple}, Multi-branch Mahalanobis \cite{anthony_use_2023}, Negative Aware Norm  \cite{park2023understanding}, Neural Activation Coverage \cite{liuneuron}, Neural Mean Discrepancy \cite{dong2022neural}, NuSA \cite{cook2020outlier}, PCX \cite{dreyer2024understanding}, Residual \cite{ndiour2020out}, TAPUUD \cite{dua2023task} and XOOD-M \cite{berglind2022xood}. Additionally, we evaluated five internal ad-hoc methods: Bayes by Backprop \cite{blundell2015weight}, CIDER \cite{ming2022exploit}, Outlier Exposure \cite{hendrycksdeep}, Reject Class \cite{chow2003optimum} and Rotation Prediction \cite{hendrycks2019using}. For methods requiring auxiliary OOD training data (Outlier Exposure and Reject Class), we used CIFAR-10 \cite{krizhevsky2009learning}. Finally, we evaluated five external OOD detection methods: two reconstruction-based  (DDPM-MSE and DDPM-LPIPS \cite{graham2023denoising}), one density-based (RealNVP \cite{kirichenko2020normalizing}) and two classifier-based (Deep SVDD \cite{ruff2018deep} and FPI \cite{tan2022detecting}). Foundation-model approaches (e.g. CLIP) were excluded from this study to prevent data leakage as large-scale pre-training likely includes these public datasets or visually similar artefacts. Although foundation models have been shown to exhibit performance drops and bias under distribution shift \cite{wang2024sober,li2023robustness}, including them would introduce confounding pre-training effects that obscure the underlying mechanism we aim to isolate. Further method details are provided in the supplementary material.



\subsection{Subspace Attribution Analysis for Mechanistic Insight}\label{Sec:subspace}
We demonstrate later in the paper that feature-based methods (which utilise the primary model's latent space) on average have a greater OOD detection drop between similar and dissimilar artefacts. We hypothesise that these large drops occur because colour variation aligns with high-variance directions in the primary model’s latent space. This can impact OOD detection as many feature-based methods downweight high-variance directions (e.g. Mahalanobis), so colour shifts that project onto high-variance directions may be under-penalised, explaining why highly dissimilar artefacts can be harder to detect than more similar ones.

We define a \emph{nuisance subspace} as the set of high-variance feature directions along which OOD artefacts (unseen during training) can induce large variation in the latent space. Let $H_{\mathrm{ID}} \in \mathbb{R}^{N \times C \times H \times W}$ denote the latent tensor  extracted from a given hidden layer in the primary model for $N$ training data inputs. We applied global average pooling over the spatial dimensions, resulting in $F_{\mathrm{ID}} \in \mathbb{R}^{N \times C}$, before computing PCA decomposition \cite{jolliffe2011principal},
\begin{equation}
    F_{\mathrm{ID}} = Z V^\top,
\end{equation}
where $Z$ are PCA coefficients, $V = [v_1,\dots,v_D] \in \mathbb{R}^{C \times C}$ are orthogonal principal directions and  
$\{\lambda_k\}_{k=1}^C$ are the corresponding variances (eigenvalues).

To identify nuisance subspaces, we projected features from similar and dissimilar OOD onto each PC. For each component, we then computed its ability to discriminate between similar and dissimilar artefacts,
\begin{equation}
I_k = \max\!\left( \mathrm{AUC}(F_{\mathrm{Diss.}} v_k,\; 
                      F_{\mathrm{Sim.}} v_k),\;
                      1 - \mathrm{AUC}(\cdot) \right),
\end{equation}
where $I_k \in [0.5,1]$ measures how strongly each PC encodes the OOD artefact's colour variation, using similar-dissimilar labels as ground truth. 

We then quantified whether nuisance variation lies in high-variance directions by computing the Spearman rank correlation \cite{spearman1961proof}, 
$\rho = \text{Spearman}(\log {\lambda_k},I_K)$,
where a positive correlation indicates that colour variation in the artefacts lies primarily along high-variance directions.

\subsection{Mitigation Strategies}
We analysed the effectiveness of two mitigation strategies for the Invisible Gorilla Effect. The first was PyTorch's colour jitter augmentation during training. We tested both a light jitter (brightness/contrast/saturation = 0.2) and a heavy jitter (brightness/contrast/saturation = 0.8) to assess the impact of augmentation strength on the effect. Inspired by Sec \ref{Sec:subspace}, the second mitigation strategy used the nuisance subspace, $U$, defined by the span of k=5 PCs (ablation in supplementary) with the highest $I_k$ scores. We projected features onto space orthogonal to U
\begin{equation}
    F_\perp = (I - U U^\top)F.
\end{equation}
We then applied feature-based methods to the projected features. We assessed mitigation by computing the nuisance subspace on the ISIC colour-chart benchmark and applying the projection to evaluate the ISIC ink benchmark (Sec 4.3).


\section{Experimental Results}
\label{sec:results}

\subsection{Evaluating the Impact of Model's ROI}
\begin{figure} 
    \centering
    \includegraphics[width=\linewidth]{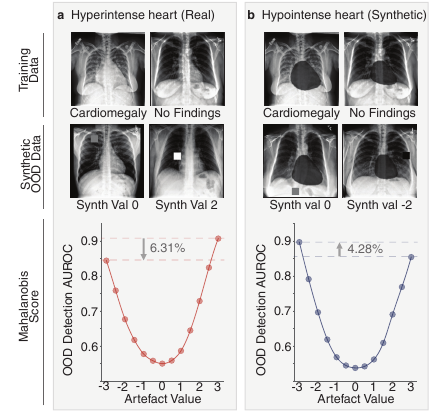} 
    \caption{Panel (a) shows training data with hyperintense hearts, (b) shows counterfactuals with hypointense hearts. Top rows show training examples and middle rows show OOD samples with synthetic squares of varying intensities. Bottom plots display AUROC for Mahalanobis Score across square intensities, with each point averaged over 25 runs on ResNet18 primary models and connected by a cubic spline. Results show that when models are trained on hyperintense hearts, this improves the OOD detection performance for hyperintense artefacts, and conversely for hypointense hearts. }
    \label{fig:Chex}
\end{figure}

To assess how a model's region of interest influences OOD detection, we compared models trained on original CheXpert images with models trained on counterfactuals where the heart was made hypointense. We then added synthetic covariate OOD artefacts (square patches occupying 1\% of the image) of varying intensity and evaluated Mahalanobis OOD detection using the best-performing feature layer (Figure~\ref{fig:Chex}). On the original (hyperintense) training set, we observed that OOD detection performance was higher for hyperintense artefacts: AUROC dropped by an average of 6.31\% when the square intensity shifted from +3 to –3. In contrast, for models trained on hypointense hearts, the trend reversed: performance favoured hypointense artefacts, with an average AUROC drop of 4.28\% when intensity shifted from –3 to +3. These findings highlight an instance of the Invisible Gorilla Effect, where the OOD detection performance appears to depend on its visual similarity (e.g. intensity) to the ROI.

\begin{table*}[ht!]
\caption{OOD detection AUROC (\%) results for the ISIC benchmark, using a ResNet18 primary model for internal methods. Detection was evaluated on artefacts that are either visually similar to skin lesions (red for ink artefacts; red, orange, yellow for colour charts) or visually dissimilar (green, purple, black for ink artefacts; green, blue, black, grey for colour charts). Results on colour-swapped counterfactuals for colour chart artefacts are also reported. Mean AUROC drop between similar and dissimilar OOD (pp) on the two real datasets are reported. Methods are grouped into four groups: feature-based, confidence-based, ad-hoc and external. 
Each entry shows the best-performing hyperparameter setting, reported as the mean AUROC over 25 seeds, with 95\% confidence intervals in brackets (in AUROC scale, not \%). 
}
\centering
\footnotesize 
\begin{tabular}{lc >{\columncolor[gray]{0.93}}c c >{\columncolor[gray]{0.93}}c c >{\columncolor[gray]{0.93}}c c >{\columncolor[gray]{0.93}}c}
\toprule
  \multirow{2}{*}{\textbf{OOD Method}} & \multicolumn{2}{c}{\textbf{Ink Artefacts}} & \multicolumn{4}{c}{\textbf{Colour Chart Artefacts}} \\ \cmidrule(lr){2-3}\cmidrule(lr){4-7}
  & Similar & Dissimilar  & Similar & Dissimilar & Synth Similar & Synth Dissimilar & $\Delta$(sim. - diss.)  \\ 
\midrule
\multicolumn{2}{l}{\textbf{Feature-based Methods }}   &  &   &   &   &  \\
CoP & 71.74 (0.02) & 65.76 (0.01) & 87.37 (0.01) & 84.35 (0.01) & 88.58 (0.01) & 83.97 (0.01) & 4.50 \\
CoRP & 71.16 (0.02) & 64.98 (0.01) & 85.75 (0.01) & 82.53 (0.01) & 84.75 (0.02) & 83.97 (0.01) & 4.70\\
FeatureNorm & 75.12 (0.01) & 52.91 (0.02) & 62.40 (0.08) & 58.14 (0.12) & 61.47 (0.02) & 46.50 (0.05) & 13.24 \\
GRAM & 80.32 (0.03) & 72.71 (0.02) & 59.08 (0.01) & 56.79 (0.01) & 61.08 (0.04) & 57.45 (0.02) & 4.95 \\
KDE (Gaussian) & 85.55 (0.01) & 68.16 (0.01) & 90.54 (0.01) & 88.97 (0.01) & 89.94 (0.01) & 86.64 (0.01) & 9.48 \\
KNN & 85.68 (0.01) & 70.10 (0.01) & 91.33 (0.03) & 90.61 (0.02) & 88.95 (0.11) & 85.91 (0.02) & 8.15 \\
LOF & 82.35 (0.01) & 62.90 (0.01) & 93.28 (0.01) & 91.37 (0.01) & 91.54 (0.01) & 89.06 (0.01) & 10.68 \\
Mahalanobis & 76.98 (0.01) & 63.64 (0.01) & 96.71 (0.01) & 95.40 (0.01) &  96.11 (0.01) & 93.77 (0.01) & 7.33 \\
MBM & 77.40 (0.01) & 63.71 (0.01) & 97.00 (0.01) & 95.52 (0.01) & 96.31 (0.01) & 94.48 (0.01) & 7.59 \\
NAN & 75.58 (0.01) & 48.46 (0.03) & 72.46 (0.05) & 68.09 (0.05) & 71.62 (0.03) & 69.09 (0.04) & 15.75 \\
NAC & 39.62 (0.02) & 37.32 (0.01) & 53.91 (0.02) & 53.03 (0.02) & 49.93 (0.02) & 46.91 (0.02) & 1.59 \\
NMD & 79.31 (0.02) & 73.73 (0.04) & 79.72 (0.01) & 78.77 (0.01) & 79.65 (0.02) & 75.98 (0.01) & 3.27 \\
NuSA & 75.02 (0.05) & 74.97 (0.05) & 56.17 (0.05) & 53.15 (0.05) & 61.92 (0.03) & 58.18 (0.02) & 1.54 \\
PCX & 75.61 (0.01) & 64.75 (0.01) & 95.16 (0.01) & 94.50 (0.01) & 92.90 (0.06) & 88.66 (0.02) & 5.76 \\
Residual & 66.00 (0.01) & 58.27 (0.01) & 93.09 (0.01) & 92.55 (0.02) & 69.86 (0.01) & 67.17 (0.01) & 4.14 \\
TAPUUD & 70.79 (0.02) & 57.01 (0.01) & 95.13 (0.01) & 93.31 (0.01) & 93.26 (0.02) & 90.34 (0.01) & 7.80 \\
XOOD-M & 80.76 (0.04) & 63.91 (0.01) & 88.50 (0.02) & 84.81 (0.03) & 90.54 (0.02) & 87.45 (0.02) & 10.27 \\

\midrule
\multicolumn{2}{l}{\textbf{Confidence-based Methods }}   &  &   &   &   &  \\
ASH & 72.41 (0.02) & 73.02 (0.01) & 75.29 (0.02) & 73.45 (0.02) & 76.27 (0.02) & 75.46 (0.02) & 0.62 \\
Deep Ensemble & 73.06 (0.01) & 72.42 (0.01) & 61.24 (0.02) & 59.15 (0.01) & 60.45 (0.01) & 58.92 (0.01) & 1.37 \\
DICE & 69.04 (0.01) & 71.30 (0.02) & 59.11 (0.03) & 57.01 (0.03) & 74.34 (0.02) & 72.00 (0.03) & -0.08 \\
GAIA-A & 68.93 (0.02) & 65.55 (0.01) & 50.59 (0.01) & 47.67 (0.01) & 50.52 (0.01) & 46.72 (0.01) & 3.15 \\
GradNorm & 75.63 (0.03) & 72.43  (0.02) & 75.11 (0.02) & 71.37 (0.01) & 70.84 (0.03) & 68.89 (0.04) & 3.47 \\
GradOrth & 72.80 (0.01) & 72.74 (0.01) & 59.80 (0.02) & 57.02 (0.02) & 63.72 (0.02) & 57.06 (0.02) & 1.42 \\
MCP & 69.83 (0.02) & 68.74 (0.02) & 57.45 (0.02) & 55.37 (0.02) & 61.72 (0.02) & 55.27 (0.02) & 1.59 \\
MC-Dropout & 70.42 (0.02) & 69.21 (0.01) & 59.33 (0.01) & 56.75 (0.02) & 58.33 (0.03) & 56.00 (0.02) & 1.90 \\
ODIN & 72.76 (0.01) & 72.36 (0.01) & 59.74 (0.02) & 57.02 (0.02) & 62.32 (0.03) & 56.87 (0.02) & 1.56\\
ReAct & 64.17 (0.05) & 59.50 (0.04) & 51.42 (0.05) & 51.30 (0.04) & 73.90 (0.02) & 71.23 (0.02) & 2.40 \\
SHE & 72.20 (0.01) & 72.36 (0.01) & 58.36 (0.02) & 55.74 (0.02) & 62.79 (0.02) & 55.75 (0.02) & 1.23 \\
ViM & 75.29 (0.01) & 74.52 (0.01) & 65.19 (0.02) & 64.04 (0.01) & 64.80 (0.02) & 56.77 (0.01) & 0.96 \\
WeiPer & 74.59 (0.04) & 73.77 (0.03) & 75.71 (0.03) & 76.53 (0.03) & 71.21 (0.03) & 65.77 (0.02) & 0.01 \\
\midrule
\textbf{Ad-hoc Methods}  & &  &   &   &   &  \\
Bayes By Backprop & 56.11 (0.05) & 54.31 (0.03) & 63.50 (0.04) & 61.78 (0.04) & 58.11 (0.05) & 56.60 (0.04) & 1.76 \\
CIDER & 72.96  (0.07) & 48.31 (0.06) & 66.39 (0.05) & 61.59 (0.03) & 67.21 (0.03) & 60.46 (0.02) & 14.73 \\
Outlier Exposure & 70.58 (0.04) & 67.29 (0.04) & 70.30 (0.02) & 67.89 (0.03) & 69.80 (0.03) & 66.55 (0.02) & 2.85 \\
Reject Class & 83.35 (0.06) & 68.33 (0.05) & 72.99 (0.04) & 65.15 (0.03) & 73.72 (0.04) & 65.46 (0.03) & 11.43 \\
Rotation Pred & 56.33 (0.04) & 52.17 (0.03) & 66.04 (0.04) & 67.15 (0.04) & 64.08 (0.03) & 64.24 (0.04) & 1.53 \\
\midrule
\textbf{External Methods }  & &  &   &   &   &  \\
Deep SVDD & 62.17 (0.01) & 39.11 (0.02) & 88.26 (0.03) & 85.13 (0.02) & 89.46 (0.02) & 83.57 (0.03) & 13.10 \\
DDPM-LPIPS & 65.83 (0.04) & 68.23 (0.04) & 87.68 (0.05) & 86.43 (0.05) & 87.92 (0.06) & 85.44 (0.04) & -0.58 \\
DDPM-MSE & 68.64 (0.06) & 70.85 (0.03) & 74.42 (0.03) & 76.85 (0.02) & 75.12 (0.04) & 75.02 (0.03) & -2.32 \\
FPI & 78.42 (0.02) & 53.65 (0.02) & 76.86 (0.02) & 73.83 (0.03) & 77.62 (0.02) & 72.80 (0.01) & 13.90 \\
RealNVP & 83.96 (0.01) & 65.62 (0.01) & 96.07 (0.01) & 94.17 (0.01) & 97.48 (0.01) & 93.02 (0.01) & 10.12 \\

\bottomrule
\end{tabular}

 \label{Tbl:IGE_results}
\end{table*}

\subsection{Large-Scale Evaluation of the Invisible Gorilla Effect}\label{Sec:IGE}
\begin{figure} 
    \centering
    \includegraphics[width=\linewidth]{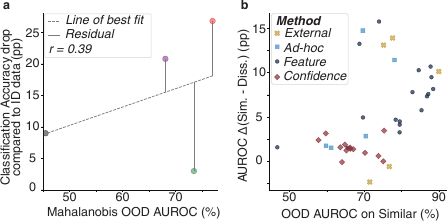} 
    \caption{a) Classification accuracy drop between ID and OOD datasets versus Mahalanobis OOD AUROC for each colour (marker colour = artefact colour) for the ISIC ink benchmark averaged over 25 ResNet18s, showing a weak correlation ($r=0.39$). b) OOD AUROC drop between similar and dissimilar colours versus OOD AUROC on similar colours across 40 methods (averaged across ISIC benchmarks), showing a weak positive correlation.}
    \label{fig:generalisation}
\end{figure}

To assess the generality of the Invisible Gorilla Effect, we performed a large-scale evaluation on real-world artefacts from the ISIC and MVTec datasets using 40 OOD detection methods. For ISIC, OOD test sets included ink annotations, colour charts and their colour-swapped counterfactuals, which were categorised as either visually similar or dissimilar to skin lesions. Table~\ref{Tbl:IGE_results} presents ISIC AUROC scores for all 40 methods, using a ResNet18 primary model for the internal methods. Our results show consistent performance drops across most methods when OOD artefacts are visually dissimilar to the model's ROI, providing empirical support for the Invisible Gorilla Effect. Analogous results for VGG16 and ViT-B/32 architectures are provided in the supplementary material, confirming that the effect holds across diverse architectures. Additional reporting of variance across hyperparameter choices is provided in the supplementary material. From Table~\ref{Tbl:IGE_results}, we observe that feature-based methods exhibit a larger mean AUROC drop ($7.1 (\mu) \pm 1.8 (\sigma)$ pp) compared to confidence-based methods ($1.5 \pm 1.1$ pp) on the real benchmarks, with means computed within each group. Greater variance in the drop in AUROC was observed for ad-hoc ($6.5 \pm 6.2$ pp) and external methods ($6.8 \pm 7.7$ pp), likely due to differences in their training regimes and architectures. Notably, the diffusion-based method DDPM-MSE was the only method that did not exhibit an AUROC drop across all three ISIC benchmarks. Overall, feature-based methods exhibited the largest drop in mean AUROC. NAC and NuSA were exceptions to this pattern, although NAC performed poorly across all tasks. We plotted the drop in OOD detection performance against performance on similar-colour artefacts (Figure~\ref{fig:generalisation}b), finding a weak correlation where higher-performing methods tended to show larger drops.


Table~\ref{Tbl:MVTec_results} presents results for the MVTec benchmarks, where we show a representative subset of 10 OOD detection methods due to space constraints. As with the ISIC experiments, we observe that performance consistently increases when OOD artefacts share visual similarity to the model's ROI. The results again show that feature-based methods typically have a larger drop in AUROC compared with confidence-based methods. Full results for all OOD detection methods, as well as analyses using VGG16 and ViT-B/32, are provided in the supplementary material.

\begin{table}[ht!]
\caption{OOD AUROC (\%) on the MVTec benchmarks using ResNet18 as the primary model. The table reports the mean AUROC over 25 seeds for each method’s best-performing hyperparameters, covering both feature and confidence-based methods.}
\centering
\footnotesize 
\begin{tabular}{lc >{\columncolor[gray]{0.93}}c c >{\columncolor[gray]{0.93}}c c }
\toprule
  \multirow{2}{*}{\textbf{Method}} & \multicolumn{2}{c}{\textbf{Pill}} & \multicolumn{2}{c}{\textbf{Metal Nut}} \\ \cmidrule(lr){2-3}\cmidrule(lr){4-5}
  & Similar & Diss.  &  Similar & Diss. & $\Delta$(S-D) \\ 
\midrule
\multicolumn{3}{l}{\textbf{Feature-based Methods }}    &     \\
CoP & 88.27 & 81.15 & 53.07 & 39.20 & 10.50 \\
KDE & 81.35 & 73.59 & 63.52 & 43.98 & 13.65 \\
KNN & 93.33 & 86.15 & 71.02 & 36.93 & 20.64 \\
Mahal. & 71.86 & 68.72 & 69.77 & 58.30 & 7.31 \\
Residual & 86.09 & 68.97 & 78.30 & 70.91 & 12.26 \\
\midrule
\multicolumn{3}{l}{\textbf{Confidence-based Methods }}    &  \\
DICE & 87.63 & 82.56 & 63.52 & 48.98 & 9.81 \\
GradNorm & 80.13 & 79.07 & 60.34 & 59.83 & 0.79 \\
MCP & 78.46 & 78.33 & 58.75 & 45.34 & 6.77 \\
ReAct & 83.72 & 80.00 & 59.43 & 45.91 & 8.62\\
WeiPer & 80.77 & 80.29 & 67.91 & 66.02 & 1.19 \\

\bottomrule
\end{tabular}

\label{Tbl:MVTec_results}
\end{table}


A potential concern is that the Invisible Gorilla Effect may simply reflect a trade-off between generalisation and OOD detection. To test this, we compared model accuracy and Mahalanobis OOD performance across ink colours in the ISIC benchmark (Fig.~\ref{fig:generalisation}a). Accuracy and OOD detection were only weakly correlated ($r = 0.39$) which is consistent with prior findings \cite{anthony_evaluating_2025, guerin_out--distribution_2023}. Large accuracy drops on dissimilar artefacts (e.g. $20$ pp for purple, $9$ pp for black) indicate that these cases remain clinically significant and should be detected. External methods (e.g. FPI, Deep SVDD) showed similar detection drops, suggesting that the Invisible Gorilla Effect is not merely a by-product of generalisation.

\subsection{Evaluation of the Mitigation Strategies}
\textbf{Colour jitter augmentation.} Table~\ref{Tbl:Augmentation} reports OOD detection results for ResNet18 models trained with light and heavy colour jitter augmentations on the ISIC ink benchmark (results for other benchmarks and architectures are provided in the Supplementary). Colour jitter shows inconsistent effects across methods: it reduces the performance gap for some (e.g. KNN) but increases it for others (e.g. CoP), and in some cases even reverses the trend by degrading detection of similar-colour artefacts (e.g. DICE). Heavier augmentations further lower ID accuracy ($-5.5$ pp), which is undesirable in deployment settings. Overall, broad colour augmentations do not reliably mitigate the Invisible Gorilla Effect and may introduce new trade-offs

\begin{table}[ht!]
\caption{OOD AUROC (\%) on the ISIC ink benchmark using 10 ResNet18 primary models. Brackets show the OOD AUROC and ID accuracy difference relative to the primary model's trained without colour jitter augmentations (Relative to Table \ref{Tbl:IGE_results}).}
\centering
\footnotesize 
\begin{tabular}{lc >{\columncolor[gray]{0.93}}c c >{\columncolor[gray]{0.93}}c  }
\toprule
  \multirow{2}{*}{\textbf{Method}} & \multicolumn{2}{c}{\textbf{Light aug. (ID acc  \textcolor{darkred}{-0.1})}} & \multicolumn{2}{c}{\textbf{Heavy aug. (ID acc \textcolor{darkred}{-5.5})}} \\ \cmidrule(lr){2-3}\cmidrule(lr){4-5}
  & Similar & Diss.  &  Similar & Diss.   \\ 
\midrule
\multicolumn{3}{l}{\textbf{Feature-based Methods }}    &     \\
CoP & 77.0 \textcolor{darkgreen}{(+5.3)} &
70.3 \textcolor{darkgreen}{(+4.6)} &
79.6 \textcolor{darkgreen}{(+7.8)} & 
69.8 \textcolor{darkgreen}{(+4.1)}  \\
KNN &  90.1 \textcolor{darkgreen}{(+4.5)} &
77.3 \textcolor{darkgreen}{(+7.2)}
& 87.9 \textcolor{darkgreen}{(+2.2)}
& 77.6 \textcolor{darkgreen}{(+7.5)} \\
NAN & 76.0 \textcolor{darkgreen}{(+0.4)} &
58.9 \textcolor{darkgreen}{(+10)} &
73.6 \textcolor{darkred}{(-2.0)} &
64.4 \textcolor{darkgreen}{(+16)}  \\
\midrule
\multicolumn{3}{l}{\textbf{Confidence-based Methods }}    &  \\
DICE & 61.8  \textcolor{darkred}{(-7.2)} & 
71.6 \textcolor{darkgreen}{(+0.3)} & 
58.1 \textcolor{darkred}{(-11)} & 
71.6 \textcolor{darkgreen}{(+0.3)} \\
ReAct & 62.8 \textcolor{darkred}{(-1.4)} & 
64.5 \textcolor{darkgreen}{(+5.0)} & 
55.6 \textcolor{darkred}{(-8.6)} & 
67.7 \textcolor{darkgreen}{(+8.2)} \\
GradNorm & 62.8 \textcolor{darkred}{(-13)} & 
71.3 \textcolor{darkgreen}{(-1.1)} & 
69.1 \textcolor{darkred}{(-6.5)} & 
75.6 \textcolor{darkgreen}{(+3.2)} \\

\bottomrule
\end{tabular}

\label{Tbl:Augmentation}
\end{table}

\textbf{Subspace projection.} We first computed the Spearman correlation between \(I_k\) and $\log$ \(\lambda_k\) for models trained on the ISIC ink benchmark (Figure~5a), finding a positive association (\(\rho = 0.47\), \(p < 1.5 \times 10^{-4}\)), indicating that OOD colour-sensitive directions align with high-variance components. This may explain the large OOD detection drops observed for feature-based methods, as methods such as Mahalanobis and FeatureNorm tend to underweight or normalise high-variance directions, reducing the penalisation of large colour shifts and causing dissimilar artefacts to be scored closer to the ID. Figure 5b reports OOD detection results for three methods for a ResNet18 primary model after projecting features to remove the nuisance subspace. Using these projected features notably reduced the gap in OOD detection performance for the methods. The trend holds across several feature-based methods, with per-method results in the supplementary.

\begin{figure}[ht]
\centering
\begin{minipage}{0.49\linewidth}
    \centering
     \subcaption{}
    \includegraphics[width=\linewidth]{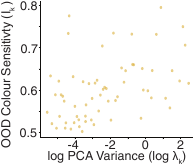}

\end{minipage}\hfill
\begin{minipage}{0.5\linewidth}
    \centering
    \subcaption{}
    \footnotesize
    \begin{tabular}{l c >{\columncolor[gray]{0.93}}c}
    \toprule
   \multirow{2}{*}{\textbf{Method}} & \multicolumn{2}{c}{\textbf{Ink Artefacts}} \\ \cmidrule(lr){2-3} 
  & Sim. & Diss. \\
    \midrule
    Mahalanobis & 77.0 & 63.6 \\
    +Proj.   & 77.5 & 75.8 \\
    \midrule
    FeatureNorm & 75.1 & 52.9 \\
    +Proj.   & 75.3 & 74.5 \\
    \midrule
    NAN & 75.6 & 48.5 \\
    +Proj.   & 75.3 & 76.8 \\
    \bottomrule
    \end{tabular}

\end{minipage}
\caption{(a) PCA variance vs.\ colour sensitivity for features from layer1.1 in a ResNet18 trained on ISIC, showing positive Spearman correlation ($\rho = 0.47, p < 1.5 \times 10^{-4}$). (b) OOD performance on ISIC ink benchmark with and without nuisance subspace projection (equation 3), averaged over 25 seeds.}
\label{Fig:eigenspace}
\end{figure}

\section{Conclusion}
This paper uncovers a previously unreported bias in OOD detection: performance improves when artefacts visually resemble the model’s region of interest, and degrades otherwise - a phenomenon we term the Invisible Gorilla Effect. Evaluating 40 methods across 7 benchmarks, we observe that this performance drop is consistent across most approaches, with feature-based methods being especially affected. We also evaluated two mitigation strategies, colour jitter augmentation and subspace projection, with the latter showing more consistent mitigation for evaluated methods.

\section{Acknowledgements}
HA is supported by a scholarship via the EPSRC Doctoral Training Partnerships programme [EP/W524311/1, EP/T517811/1]. ZL is supported by a scholarship provided by the EPSRC Doctoral Training Partnerships programme [EP/W524311/1]. HW is supported by the EPSRC Centre for Doctoral Training in Health Data Science [EP/S02428X/1]. The authors also acknowledge the use of the University of Oxford Advanced Research Computing (ARC) facility for the work (http://dx.doi.org/10.5281/zenodo.22558).

{
    \small
    \bibliographystyle{ieeenat_fullname}
    \bibliography{main}

\begin{thebibliography}{100}
\providecommand{\natexlab}[1]{#1}
\providecommand{\url}[1]{\texttt{#1}}
\expandafter\ifx\csname urlstyle\endcsname\relax
  \providecommand{\doi}[1]{doi: #1}\else
  \providecommand{\doi}{doi: \begingroup \urlstyle{rm}\Url}\fi

\bibitem[Achtibat et~al.(2023)Achtibat, Dreyer, Eisenbraun, Bosse, Wiegand, Samek, and Lapuschkin]{achtibat_attribution_2023}
Reduan Achtibat, Maximilian Dreyer, Ilona Eisenbraun, Sebastian Bosse, Thomas Wiegand, Wojciech Samek, and Sebastian Lapuschkin.
\newblock From attribution maps to human-understandable explanations through {Concept} {Relevance} {Propagation}.
\newblock \emph{Nature Machine Intelligence}, 5\penalty0 (9):\penalty0 1006--1019, 2023.
\newblock Number: 9 Publisher: Nature Publishing Group.

\bibitem[Amodei et~al.(2016)Amodei, Olah, Steinhardt, Christiano, Schulman, and Mané]{amodei_concrete_2016}
Dario Amodei, Chris Olah, Jacob Steinhardt, Paul Christiano, John Schulman, and Dan Mané.
\newblock Concrete {Problems} in {AI} {Safety}, 2016.
\newblock arXiv:1606.06565 [cs].

\bibitem[Anthony and Kamnitsas(2023)]{anthony_use_2023}
Harry Anthony and Konstantinos Kamnitsas.
\newblock On the {Use} of {Mahalanobis} {Distance} for {Out}-of-distribution {Detection} with {Neural} {Networks} for {Medical} {Imaging}.
\newblock In \emph{Uncertainty for {Safe} {Utilization} of {Machine} {Learning} in {Medical} {Imaging}}, pages 136--146, Cham, 2023. Springer Nature Switzerland.

\bibitem[Anthony and Kamnitsas(2025)]{anthony_evaluating_2025}
Harry Anthony and Konstantinos Kamnitsas.
\newblock Evaluating {Reliability} in {Medical} {DNNs}: {A} {Critical} {Analysis} of {Feature} and {Confidence}-{Based} {OOD} {Detection}.
\newblock In \emph{Uncertainty for {Safe} {Utilization} of {Machine} {Learning} in {Medical} {Imaging}}, pages 160--170, Cham, 2025. Springer Nature Switzerland.

\bibitem[Averly and Chao(2023)]{averly2023unified}
Reza Averly and Wei-Lun Chao.
\newblock Unified out-of-distribution detection: A model-specific perspective.
\newblock In \emph{Proceedings of the IEEE/CVF International Conference on Computer Vision}, pages 1453--1463, 2023.

\bibitem[Bach et~al.(2015)Bach, Binder, Montavon, Klauschen, M{\"u}ller, and Samek]{bach2015pixel}
Sebastian Bach, Alexander Binder, Gr{\'e}goire Montavon, Frederick Klauschen, Klaus-Robert M{\"u}ller, and Wojciech Samek.
\newblock On pixel-wise explanations for non-linear classifier decisions by layer-wise relevance propagation.
\newblock \emph{PloS one}, 10\penalty0 (7):\penalty0 e0130140, 2015.

\bibitem[Baptista et~al.(2025)Baptista, Dasgupta, Kovachki, Oberai, and Stuart]{baptista_memorization_2025}
Ricardo Baptista, Agnimitra Dasgupta, Nikola~B. Kovachki, Assad Oberai, and Andrew~M. Stuart.
\newblock Memorization and {Regularization} in {Generative} {Diffusion} {Models}, 2025.
\newblock arXiv:2501.15785 [cs].

\bibitem[Behpour et~al.(2023)Behpour, Doan, Li, He, Gou, and Ren]{behpour2023gradorth}
Sima Behpour, Thang~Long Doan, Xin Li, Wenbin He, Liang Gou, and Liu Ren.
\newblock Gradorth: A simple yet efficient out-of-distribution detection with orthogonal projection of gradients.
\newblock \emph{Advances in Neural Information Processing Systems}, 36:\penalty0 38206--38230, 2023.

\bibitem[Berglind et~al.(2022)Berglind, Temam, Mukhopadhyay, Das, Sajol, Kumar, and Kallurupalli]{berglind2022xood}
Frej Berglind, Haron Temam, Supratik Mukhopadhyay, Kamalika Das, Md~Saiful~Islam Sajol, Sricharan Kumar, and Kumar Kallurupalli.
\newblock Xood: Extreme value based out-of-distribution detection for image classification.
\newblock \emph{arXiv preprint arXiv:2208.00629}, 2022.

\bibitem[Bergmann et~al.(2019)Bergmann, Fauser, Sattlegger, and Steger]{bergmann2019mvtec}
Paul Bergmann, Michael Fauser, David Sattlegger, and Carsten Steger.
\newblock Mvtec ad--a comprehensive real-world dataset for unsupervised anomaly detection.
\newblock In \emph{Proceedings of the IEEE/CVF conference on computer vision and pattern recognition}, pages 9592--9600, 2019.

\bibitem[Blei et~al.(2017)Blei, Kucukelbir, and McAuliffe]{blei2017variational}
David~M Blei, Alp Kucukelbir, and Jon~D McAuliffe.
\newblock Variational inference: A review for statisticians.
\newblock \emph{Journal of the American statistical Association}, 112\penalty0 (518):\penalty0 859--877, 2017.

\bibitem[Blundell et~al.(2015)Blundell, Cornebise, Kavukcuoglu, and Wierstra]{blundell2015weight}
Charles Blundell, Julien Cornebise, Koray Kavukcuoglu, and Daan Wierstra.
\newblock Weight uncertainty in neural network.
\newblock In \emph{International conference on machine learning}, pages 1613--1622. PMLR, 2015.

\bibitem[Carpenter(2001)]{carpenter2001sights}
Siri Carpenter.
\newblock Sights unseen.
\newblock \emph{Monitor on Psychology}, 32\penalty0 (4):\penalty0 54--57, 2001.

\bibitem[Caterini and Loaiza-Ganem(2022)]{caterini2022entropic}
Anthony~L Caterini and Gabriel Loaiza-Ganem.
\newblock Entropic issues in likelihood-based ood detection.
\newblock In \emph{I (Still) Can't Believe It's Not Better! Workshop at NeurIPS 2021}, pages 21--26. PMLR, 2022.

\bibitem[Chabris and Simons(2011)]{chabris2011invisible}
Christopher Chabris and Daniel Simons.
\newblock \emph{The invisible gorilla: How our intuitions deceive us}.
\newblock Harmony, 2011.

\bibitem[Chen et~al.(2023)Chen, Li, Qu, Wang, Wan, and Xiao]{chen2023gaia}
Jinggang Chen, Junjie Li, Xiaoyang Qu, Jianzong Wang, Jiguang Wan, and Jing Xiao.
\newblock Gaia: Delving into gradient-based attribution abnormality for out-of-distribution detection.
\newblock \emph{Advances in Neural Information Processing Systems}, 36:\penalty0 79946--79958, 2023.

\bibitem[Chen et~al.(2018)Chen, Badrinarayanan, Lee, and Rabinovich]{chen2018gradnorm}
Zhao Chen, Vijay Badrinarayanan, Chen-Yu Lee, and Andrew Rabinovich.
\newblock Gradnorm: Gradient normalization for adaptive loss balancing in deep multitask networks.
\newblock In \emph{International conference on machine learning}, pages 794--803. PMLR, 2018.

\bibitem[Cheng et~al.(2025)Cheng, Zhu, Zhang, and Liu]{cheng2025average}
Zhen Cheng, Fei Zhu, Xu-Yao Zhang, and Cheng-Lin Liu.
\newblock Average of pruning: Improving performance and stability of out-of-distribution detection.
\newblock \emph{IEEE Transactions on Neural Networks and Learning Systems}, 2025.

\bibitem[Chow(2003)]{chow2003optimum}
C Chow.
\newblock On optimum recognition error and reject tradeoff.
\newblock \emph{IEEE Transactions on information theory}, 16\penalty0 (1):\penalty0 41--46, 2003.

\bibitem[Codella et~al.(2019)Codella, Rotemberg, Tschandl, Celebi, Dusza, Gutman, Helba, Kalloo, Liopyris, Marchetti, Kittler, and Halpern]{codella_skin_2019}
Noel Codella, Veronica Rotemberg, Philipp Tschandl, M.~Emre Celebi, Stephen Dusza, David Gutman, Brian Helba, Aadi Kalloo, Konstantinos Liopyris, Michael Marchetti, Harald Kittler, and Allan Halpern.
\newblock Skin {Lesion} {Analysis} {Toward} {Melanoma} {Detection} 2018: {A} {Challenge} {Hosted} by the {International} {Skin} {Imaging} {Collaboration} ({ISIC}), 2019.
\newblock arXiv:1902.03368 [cs].

\bibitem[Cook et~al.(2020)Cook, Zare, and Gader]{cook2020outlier}
Matthew Cook, Alina Zare, and Paul Gader.
\newblock Outlier detection through null space analysis of neural networks.
\newblock \emph{arXiv preprint arXiv:2007.01263}, 2020.

\bibitem[Dhamija et~al.(2018)Dhamija, G{\"u}nther, and Boult]{dhamija2018reducing}
Akshay~Raj Dhamija, Manuel G{\"u}nther, and Terrance Boult.
\newblock Reducing network agnostophobia.
\newblock \emph{Advances in Neural Information Processing Systems}, 31, 2018.

\bibitem[Dinh et~al.(2017)Dinh, Sohl-Dickstein, and Bengio]{dinh2017density}
Laurent Dinh, Jascha Sohl-Dickstein, and Samy Bengio.
\newblock Density estimation using real nvp.
\newblock In \emph{International Conference on Learning Representations}, 2017.

\bibitem[Djurisic et~al.(2022)Djurisic, Bozanic, Ashok, and Liu]{djurisicextremely}
Andrija Djurisic, Nebojsa Bozanic, Arjun Ashok, and Rosanne Liu.
\newblock Extremely simple activation shaping for out-of-distribution detection.
\newblock In \emph{The Eleventh International Conference on Learning Representations}, 2022.

\bibitem[Dong et~al.(2022)Dong, Guo, Li, Ting, Liu, and Kung]{dong2022neural}
Xin Dong, Junfeng Guo, Ang Li, Wei-Te Ting, Cong Liu, and HT Kung.
\newblock Neural mean discrepancy for efficient out-of-distribution detection.
\newblock In \emph{Proceedings of the IEEE/CVF Conference on Computer Vision and Pattern Recognition}, pages 19217--19227, 2022.

\bibitem[Dosovitskiy et~al.(2021)Dosovitskiy, Beyer, Kolesnikov, Weissenborn, Zhai, Unterthiner, Dehghani, Minderer, Heigold, Gelly, Uszkoreit, and Houlsby]{dosovitskiy_image_2021}
Alexey Dosovitskiy, Lucas Beyer, Alexander Kolesnikov, Dirk Weissenborn, Xiaohua Zhai, Thomas Unterthiner, Mostafa Dehghani, Matthias Minderer, Georg Heigold, Sylvain Gelly, Jakob Uszkoreit, and Neil Houlsby.
\newblock An {Image} is {Worth} 16x16 {Words}: {Transformers} for {Image} {Recognition} at {Scale}, 2021.
\newblock arXiv:2010.11929 [cs].

\bibitem[Draelos and Carin(2020)]{draelos2020use}
Rachel~Lea Draelos and Lawrence Carin.
\newblock Use hirescam instead of grad-cam for faithful explanations of convolutional neural networks.
\newblock \emph{arXiv preprint arXiv:2011.08891}, 2020.

\bibitem[Dreyer et~al.(2024)Dreyer, Achtibat, Samek, and Lapuschkin]{dreyer2024understanding}
Maximilian Dreyer, Reduan Achtibat, Wojciech Samek, and Sebastian Lapuschkin.
\newblock Understanding the (extra-) ordinary: Validating deep model decisions with prototypical concept-based explanations.
\newblock In \emph{2024 IEEE/CVF Conference on Computer Vision and Pattern Recognition Workshops (CVPRW)}, pages 3491--3501. IEEE, 2024.

\bibitem[Du et~al.(2024)Du, Fang, Diakonikolas, and Li]{du2024does}
Xuefeng Du, Zhen Fang, Ilias Diakonikolas, and Yixuan Li.
\newblock How does unlabeled data provably help out-of-distribution detection?
\newblock \emph{arXiv preprint arXiv:2402.03502}, 2024.

\bibitem[Dua et~al.(2023)Dua, Yang, Li, and Choi]{dua2023task}
Radhika Dua, Seongjun Yang, Yixuan Li, and Edward Choi.
\newblock Task agnostic and post-hoc unseen distribution detection.
\newblock In \emph{Proceedings of the IEEE/CVF Winter Conference on Applications of Computer Vision}, pages 1350--1359, 2023.

\bibitem[Erdil et~al.(2021)Erdil, Chaitanya, Karani, and Konukoglu]{erdil2021task}
Ertunc Erdil, Krishna Chaitanya, Neerav Karani, and Ender Konukoglu.
\newblock Task-agnostic out-of-distribution detection using kernel density estimation.
\newblock In \emph{International Workshop on Uncertainty for Safe Utilization of Machine Learning in Medical Imaging}, pages 91--101. Springer, 2021.

\bibitem[Erion et~al.(2021)Erion, Janizek, Sturmfels, Lundberg, and Lee]{erion2021improving}
Gabriel Erion, Joseph~D Janizek, Pascal Sturmfels, Scott~M Lundberg, and Su-In Lee.
\newblock Improving performance of deep learning models with axiomatic attribution priors and expected gradients.
\newblock \emph{Nature machine intelligence}, 3\penalty0 (7):\penalty0 620--631, 2021.

\bibitem[Esteva et~al.(2017)Esteva, Kuprel, Novoa, Ko, Swetter, Blau, and Thrun]{esteva_dermatologist-level_2017}
Andre Esteva, Brett Kuprel, Roberto~A. Novoa, Justin Ko, Susan~M. Swetter, Helen~M. Blau, and Sebastian Thrun.
\newblock Dermatologist-level classification of skin cancer with deep neural networks.
\newblock \emph{Nature}, 542\penalty0 (7639):\penalty0 115--118, 2017.
\newblock Number: 7639 Publisher: Nature Publishing Group.

\bibitem[{European Commission}(2021)]{eu2021aiact}
{European Commission}.
\newblock Proposal for a regulation of the european parliament and the council laying down harmonised rules on artificial intelligence (artificial intelligence act) and amending certain union legislative acts, 2021.
\newblock Accessed: 2025-07-25.

\bibitem[Fang et~al.(2024)Fang, Tao, Lv, He, Huang, and Yang]{fang2024kernel}
Kun Fang, Qinghua Tao, Kexin Lv, Mingzhen He, Xiaolin Huang, and Jie Yang.
\newblock Kernel pca for out-of-distribution detection.
\newblock \emph{Advances in Neural Information Processing Systems}, 37:\penalty0 134317--134344, 2024.

\bibitem[Fang et~al.(2022)Fang, Li, Lu, Dong, Han, and Liu]{fang2022out}
Zhen Fang, Yixuan Li, Jie Lu, Jiahua Dong, Bo Han, and Feng Liu.
\newblock Is out-of-distribution detection learnable?
\newblock \emph{Advances in Neural Information Processing Systems}, 35:\penalty0 37199--37213, 2022.

\bibitem[Gaggion et~al.(2024)Gaggion, Mosquera, Mansilla, Saidman, Aineseder, Milone, and Ferrante]{gaggion2024chexmask}
Nicol{\'a}s Gaggion, Candelaria Mosquera, Lucas Mansilla, Julia~Mariel Saidman, Martina Aineseder, Diego~H Milone, and Enzo Ferrante.
\newblock Chexmask: a large-scale dataset of anatomical segmentation masks for multi-center chest x-ray images.
\newblock \emph{Scientific Data}, 11\penalty0 (1):\penalty0 511, 2024.

\bibitem[Gal and Ghahramani(2016)]{gal2016dropout}
Yarin Gal and Zoubin Ghahramani.
\newblock Dropout as a bayesian approximation: Representing model uncertainty in deep learning.
\newblock In \emph{international conference on machine learning}, pages 1050--1059. PMLR, 2016.

\bibitem[Gao et~al.(2024)Gao, Chen, Xiang, and Xu]{gao2024comprehensive}
Junyu Gao, Mengyuan Chen, Liangyu Xiang, and Changsheng Xu.
\newblock A comprehensive survey on evidential deep learning and its applications.
\newblock \emph{arXiv preprint arXiv:2409.04720}, 2024.

\bibitem[Gonz{\'a}lez et~al.(2024)Gonz{\'a}lez, Fuchs, dos Santos, Matthies, Trenz, Gr{\"u}ning, Chaudhari, Larson, Othman, Kim, et~al.]{gonzalez2024regulating}
Camila Gonz{\'a}lez, Moritz Fuchs, Daniel~Pinto dos Santos, Philipp Matthies, Manuel Trenz, Maximilian Gr{\"u}ning, Akshay Chaudhari, David~B Larson, Ahmed~E Othman, Moon Kim, et~al.
\newblock Regulating radiology ai medical devices that evolve in their lifecycle.
\newblock \emph{CoRR}, 2024.

\bibitem[Graham et~al.(2023)Graham, Pinaya, Tudosiu, Nachev, Ourselin, and Cardoso]{graham2023denoising}
Mark~S Graham, Walter~HL Pinaya, Petru-Daniel Tudosiu, Parashkev Nachev, Sebastien Ourselin, and Jorge Cardoso.
\newblock Denoising diffusion models for out-of-distribution detection.
\newblock In \emph{Proceedings of the IEEE/CVF Conference on Computer Vision and Pattern Recognition}, pages 2948--2957, 2023.

\bibitem[Granz et~al.(2024)Granz, Heurich, and Landgraf]{granz2024weiper}
Maximilian Granz, Manuel Heurich, and Tim Landgraf.
\newblock Weiper: Ood detection using weight perturbations of class projections.
\newblock \emph{Advances in Neural Information Processing Systems}, 37:\penalty0 35879--35908, 2024.

\bibitem[Gulshan et~al.(2019)Gulshan, Rajan, Widner, Wu, Wubbels, Rhodes, Whitehouse, Coram, Corrado, Ramasamy, Raman, Peng, and Webster]{gulshan_performance_2019}
Varun Gulshan, Renu~P. Rajan, Kasumi Widner, Derek Wu, Peter Wubbels, Tyler Rhodes, Kira Whitehouse, Marc Coram, Greg Corrado, Kim Ramasamy, Rajiv Raman, Lily Peng, and Dale~R. Webster.
\newblock Performance of a {Deep}-{Learning} {Algorithm} vs {Manual} {Grading} for {Detecting} {Diabetic} {Retinopathy} in {India}.
\newblock \emph{JAMA Ophthalmology}, 137\penalty0 (9):\penalty0 987--993, 2019.

\bibitem[Gutbrod et~al.(2025)Gutbrod, Rauber, Nunes, and Palm]{gutbrod2025openmibood}
Max Gutbrod, David Rauber, Danilo~Weber Nunes, and Christoph Palm.
\newblock Openmibood: Open medical imaging benchmarks for out-of-distribution detection.
\newblock In \emph{Proceedings of the Computer Vision and Pattern Recognition Conference}, pages 25874--25886, 2025.

\bibitem[Guérin et~al.(2023)Guérin, Delmas, Ferreira, and Guiochet]{guerin_out--distribution_2023}
Joris Guérin, Kevin Delmas, Raul~Sena Ferreira, and Jérémie Guiochet.
\newblock Out-{Of}-{Distribution} {Detection} {Is} {Not} {All} {You} {Need}, 2023.
\newblock arXiv:2211.16158 [cs, eess].

\bibitem[He et~al.(2016)He, Zhang, Ren, and Sun]{he_deep_2016}
Kaiming He, Xiangyu Zhang, Shaoqing Ren, and Jian Sun.
\newblock Deep {Residual} {Learning} for {Image} {Recognition}.
\newblock pages 770--778, 2016.

\bibitem[Hendrycks and Gimpel(2016)]{hendrycks_baseline_2016}
Dan Hendrycks and Kevin Gimpel.
\newblock A {Baseline} for {Detecting} {Misclassified} and {Out}-of-{Distribution} {Examples} in {Neural} {Networks}.
\newblock 2016.

\bibitem[Hendrycks et~al.(2018)Hendrycks, Mazeika, and Dietterich]{hendrycksdeep}
Dan Hendrycks, Mantas Mazeika, and Thomas Dietterich.
\newblock Deep anomaly detection with outlier exposure.
\newblock In \emph{International Conference on Learning Representations}, 2018.

\bibitem[Hendrycks et~al.(2019)Hendrycks, Mazeika, Kadavath, and Song]{hendrycks2019using}
Dan Hendrycks, Mantas Mazeika, Saurav Kadavath, and Dawn Song.
\newblock Using self-supervised learning can improve model robustness and uncertainty.
\newblock \emph{Advances in neural information processing systems}, 32, 2019.

\bibitem[Irvin et~al.(2019)Irvin, Rajpurkar, Ko, Yu, Ciurea-Ilcus, Chute, Marklund, Haghgoo, Ball, Shpanskaya, Seekins, Mong, Halabi, Sandberg, Jones, Larson, Langlotz, Patel, Lungren, and Ng]{irvin_chexpert_2019}
Jeremy Irvin, Pranav Rajpurkar, Michael Ko, Yifan Yu, Silviana Ciurea-Ilcus, Chris Chute, Henrik Marklund, Behzad Haghgoo, Robyn Ball, Katie Shpanskaya, Jayne Seekins, David~A. Mong, Safwan~S. Halabi, Jesse~K. Sandberg, Ricky Jones, David~B. Larson, Curtis~P. Langlotz, Bhavik~N. Patel, Matthew~P. Lungren, and Andrew~Y. Ng.
\newblock {CheXpert}: {A} {Large} {Chest} {Radiograph} {Dataset} with {Uncertainty} {Labels} and {Expert} {Comparison}, 2019.
\newblock arXiv:1901.07031 [cs, eess].

\bibitem[Janai et~al.(2020)Janai, G{\"u}ney, Behl, Geiger, et~al.]{janai2020computer}
Joel Janai, Fatma G{\"u}ney, Aseem Behl, Andreas Geiger, et~al.
\newblock Computer vision for autonomous vehicles: Problems, datasets and state of the art.
\newblock \emph{Foundations and trends{\textregistered} in computer graphics and vision}, 12\penalty0 (1--3):\penalty0 1--308, 2020.

\bibitem[Jiang et~al.(2021)Jiang, Zhang, Hou, Cheng, and Wei]{Jiang2021LayerCAM}
Peng-Tao Jiang, Changlin Zhang, Qibin Hou, Ming-Ming Cheng, and Yunchao Wei.
\newblock Layercam: Exploring hierarchical class activation maps for localization.
\newblock \emph{IEEE Transactions on Image Processing}, 30:\penalty0 5875--5888, 2021.

\bibitem[Jolliffe(2011)]{jolliffe2011principal}
Ian Jolliffe.
\newblock Principal component analysis.
\newblock In \emph{International encyclopedia of statistical science}, pages 1094--1096. Springer, 2011.

\bibitem[Kamkari et~al.(2024)Kamkari, Ross, Cresswell, Caterini, Krishnan, and Loaiza-Ganem]{kamkari2024geometric}
Hamidreza Kamkari, Brendan~Leigh Ross, Jesse~C Cresswell, Anthony~L Caterini, Rahul Krishnan, and Gabriel Loaiza-Ganem.
\newblock A geometric explanation of the likelihood ood detection paradox.
\newblock In \emph{International Conference on Machine Learning}, pages 22908--22935. PMLR, 2024.

\bibitem[Kirichenko et~al.(2020)Kirichenko, Izmailov, and Wilson]{kirichenko2020normalizing}
Polina Kirichenko, Pavel Izmailov, and Andrew~G Wilson.
\newblock Why normalizing flows fail to detect out-of-distribution data.
\newblock \emph{Advances in neural information processing systems}, 33:\penalty0 20578--20589, 2020.

\bibitem[Kirillov et~al.(2023)Kirillov, Mintun, Ravi, Mao, Rolland, Gustafson, Xiao, Whitehead, Berg, Lo, Dollar, and Girshick]{kirillov_segment_2023}
Alexander Kirillov, Eric Mintun, Nikhila Ravi, Hanzi Mao, Chloe Rolland, Laura Gustafson, Tete Xiao, Spencer Whitehead, Alexander~C. Berg, Wan-Yen Lo, Piotr Dollar, and Ross Girshick.
\newblock Segment {Anything}.
\newblock pages 4015--4026, 2023.

\bibitem[Krizhevsky and Hinton(2009)]{krizhevsky2009learning}
Alex Krizhevsky and Geoffrey Hinton.
\newblock Learning multiple layers of features from tiny images.
\newblock Technical report, University of Toronto, 2009.

\bibitem[Lakshminarayanan et~al.(2017)Lakshminarayanan, Pritzel, and Blundell]{lakshminarayanan_simple_2017}
Balaji Lakshminarayanan, Alexander Pritzel, and Charles Blundell.
\newblock Simple and {Scalable} {Predictive} {Uncertainty} {Estimation} using {Deep} {Ensembles}.
\newblock In \emph{Advances in {Neural} {Information} {Processing} {Systems}}. Curran Associates, Inc., 2017.

\bibitem[Lambert et~al.(2023)Lambert, Forbes, Doyle, and Dojat]{lambert2023multi}
Benjamin Lambert, Florence Forbes, Senan Doyle, and Michel Dojat.
\newblock Multi-layer aggregation as a key to feature-based ood detection.
\newblock In \emph{International workshop on uncertainty for safe utilization of machine learning in medical imaging}, pages 104--114. Springer, 2023.

\bibitem[Lee et~al.(2018)Lee, Lee, Lee, and Shin]{lee2018simple}
Kimin Lee, Kibok Lee, Honglak Lee, and Jinwoo Shin.
\newblock A simple unified framework for detecting out-of-distribution samples and adversarial attacks.
\newblock \emph{Advances in neural information processing systems}, 31, 2018.

\bibitem[Li et~al.(2023)Li, Zhang, Hong, Li, and Zhang]{li2023robustness}
Kaican Li, Yifan Zhang, Lanqing Hong, Zhenguo Li, and Nevin~L Zhang.
\newblock Robustness may be more brittle than we think under different degrees of distribution shifts.
\newblock \emph{arXiv preprint arXiv:2310.06622}, 2023.

\bibitem[Liang et~al.(2018)Liang, Li, and Srikant]{liang2018enhancing}
Shiyu Liang, Yixuan Li, and R Srikant.
\newblock Enhancing the reliability of out-of-distribution image detection in neural networks.
\newblock In \emph{International Conference on Learning Representations}, 2018.

\bibitem[Liu et~al.(2023)Liu, Chris, Li, Ma, and Wang]{liuneuron}
Yibing Liu, XING Chris, Haoliang Li, Lei Ma, and Shiqi Wang.
\newblock Neuron activation coverage: Rethinking out-of-distribution detection and generalization.
\newblock In \emph{The Twelfth International Conference on Learning Representations}, 2023.

\bibitem[Liu et~al.(2024)Liu, Tian, Li, Ma, and Wang]{liu2024neuron}
Yibing Liu, Chris~Xing Tian, Haoliang Li, Lei Ma, and Shiqi Wang.
\newblock Neuron activation coverage: Rethinking out-of-distribution detection and generalization.
\newblock In \emph{12th International Conference on Learning Representations (ICLR 2024)}. International Conference on Learning Representations, ICLR, 2024.

\bibitem[Loshchilov and Hutter(2017)]{loshchilovdecoupled}
Ilya Loshchilov and Frank Hutter.
\newblock Decoupled weight decay regularization.
\newblock In \emph{International Conference on Learning Representations}, 2017.

\bibitem[Luo et~al.(2001)Luo, Cui, and Rigg]{Luo2001CIEDE2000}
M.R. Luo, G. Cui, and B. Rigg.
\newblock The development of the cie 2000 colour-difference formula: Ciede2000.
\newblock \emph{Color Research \& Application}, 26\penalty0 (5):\penalty0 340--350, 2001.

\bibitem[Ming et~al.(2022)Ming, Sun, Dia, and Li]{ming2022exploit}
Yifei Ming, Yiyou Sun, Ousmane Dia, and Yixuan Li.
\newblock How to exploit hyperspherical embeddings for out-of-distribution detection?
\newblock \emph{arXiv preprint arXiv:2203.04450}, 2022.

\bibitem[Molnar(2020)]{molnar2020interpretable}
Christoph Molnar.
\newblock \emph{Interpretable machine learning}.
\newblock Lulu. com, 2020.

\bibitem[Most et~al.(2001)Most, Simons, Scholl, Jimenez, Clifford, and Chabris]{most2001not}
Steven~B Most, Daniel~J Simons, Brian~J Scholl, Rachel Jimenez, Erin Clifford, and Christopher~F Chabris.
\newblock How not to be seen: The contribution of similarity and selective ignoring to sustained inattentional blindness.
\newblock \emph{Psychological science}, 12\penalty0 (1):\penalty0 9--17, 2001.

\bibitem[Ndiour et~al.(2020)Ndiour, Ahuja, and Tickoo]{ndiour2020out}
Ibrahima Ndiour, Nilesh Ahuja, and Omesh Tickoo.
\newblock Out-of-distribution detection with subspace techniques and probabilistic modeling of features.
\newblock \emph{arXiv preprint arXiv:2012.04250}, 2020.

\bibitem[Park et~al.(2023)Park, Chai, Yoon, and Teoh]{park2023understanding}
Jaewoo Park, Jacky Chen~Long Chai, Jaeho Yoon, and Andrew Beng~Jin Teoh.
\newblock Understanding the feature norm for out-of-distribution detection.
\newblock In \emph{Proceedings of the IEEE/CVF international conference on computer vision}, pages 1557--1567, 2023.

\bibitem[Perone et~al.(2019)Perone, Ballester, Barros, and Cohen-Adad]{perone2019unsupervised}
Christian~S Perone, Pedro Ballester, Rodrigo~C Barros, and Julien Cohen-Adad.
\newblock Unsupervised domain adaptation for medical imaging segmentation with self-ensembling.
\newblock \emph{NeuroImage}, 194:\penalty0 1--11, 2019.

\bibitem[Rafiee et~al.(2022)Rafiee, Gholamipoor, Adaloglou, Jaxy, Ramakers, and Kollmann]{rafiee2022self}
Nima Rafiee, Rahil Gholamipoor, Nikolas Adaloglou, Simon Jaxy, Julius Ramakers, and Markus Kollmann.
\newblock Self-supervised anomaly detection by self-distillation and negative sampling.
\newblock In \emph{International Conference on Artificial Neural Networks}, pages 459--470. Springer, 2022.

\bibitem[Ren et~al.(2019)Ren, Liu, Fertig, Snoek, Poplin, Depristo, Dillon, and Lakshminarayanan]{ren2019likelihood}
Jie Ren, Peter~J Liu, Emily Fertig, Jasper Snoek, Ryan Poplin, Mark Depristo, Joshua Dillon, and Balaji Lakshminarayanan.
\newblock Likelihood ratios for out-of-distribution detection.
\newblock \emph{Advances in neural information processing systems}, 32, 2019.

\bibitem[Ren et~al.(2021)Ren, Fort, Liu, Roy, Padhy, and Lakshminarayanan]{ren2021simple}
Jie Ren, Stanislav Fort, Jeremiah Liu, Abhijit~Guha Roy, Shreyas Padhy, and Balaji Lakshminarayanan.
\newblock A simple fix to mahalanobis distance for improving near-ood detection.
\newblock \emph{arXiv preprint arXiv:2106.09022}, 2021.

\bibitem[Ronneberger et~al.(2015)Ronneberger, Fischer, and Brox]{ronneberger2015u}
Olaf Ronneberger, Philipp Fischer, and Thomas Brox.
\newblock U-net: Convolutional networks for biomedical image segmentation.
\newblock In \emph{International Conference on Medical image computing and computer-assisted intervention}, pages 234--241. Springer, 2015.

\bibitem[Ruff et~al.(2018)Ruff, Vandermeulen, Goernitz, Deecke, Siddiqui, Binder, M{\"u}ller, and Kloft]{ruff2018deep}
Lukas Ruff, Robert Vandermeulen, Nico Goernitz, Lucas Deecke, Shoaib~Ahmed Siddiqui, Alexander Binder, Emmanuel M{\"u}ller, and Marius Kloft.
\newblock Deep one-class classification.
\newblock In \emph{International conference on machine learning}, pages 4393--4402. PMLR, 2018.

\bibitem[Sastry and Oore(2020)]{sastry2020detecting}
Chandramouli~Shama Sastry and Sageev Oore.
\newblock Detecting out-of-distribution examples with gram matrices.
\newblock In \emph{International conference on machine learning}, pages 8491--8501. PMLR, 2020.

\bibitem[Scheirer et~al.(2012)Scheirer, de~Rezende~Rocha, Sapkota, and Boult]{scheirer2012toward}
Walter~J Scheirer, Anderson de Rezende~Rocha, Archana Sapkota, and Terrance~E Boult.
\newblock Toward open set recognition.
\newblock \emph{IEEE transactions on pattern analysis and machine intelligence}, 35\penalty0 (7):\penalty0 1757--1772, 2012.

\bibitem[Selvaraju et~al.(2017)Selvaraju, Cogswell, Das, Vedantam, Parikh, and Batra]{selvaraju2017grad}
Ramprasaath~R Selvaraju, Michael Cogswell, Abhishek Das, Ramakrishna Vedantam, Devi Parikh, and Dhruv Batra.
\newblock Grad-cam: Visual explanations from deep networks via gradient-based localization.
\newblock In \emph{Proceedings of the IEEE international conference on computer vision}, pages 618--626, 2017.

\bibitem[Simonyan and Zisserman(2015)]{simonyan_very_2015}
Karen Simonyan and Andrew Zisserman.
\newblock Very {Deep} {Convolutional} {Networks} for {Large}-{Scale} {Image} {Recognition}, 2015.
\newblock arXiv:1409.1556 [cs].

\bibitem[Spearman(1961)]{spearman1961proof}
Charles Spearman.
\newblock The proof and measurement of association between two things.
\newblock \emph{The American Journal of Psychology}, 1961.

\bibitem[Sun and Li(2022)]{sun2022dice}
Yiyou Sun and Yixuan Li.
\newblock Dice: Leveraging sparsification for out-of-distribution detection.
\newblock In \emph{European conference on computer vision}, pages 691--708. Springer, 2022.

\bibitem[Sun et~al.(2021)Sun, Guo, and Li]{sun2021react}
Yiyou Sun, Chuan Guo, and Yixuan Li.
\newblock React: Out-of-distribution detection with rectified activations.
\newblock \emph{Advances in neural information processing systems}, 34:\penalty0 144--157, 2021.

\bibitem[Sun et~al.(2022)Sun, Ming, Zhu, and Li]{sun2022out}
Yiyou Sun, Yifei Ming, Xiaojin Zhu, and Yixuan Li.
\newblock Out-of-distribution detection with deep nearest neighbors.
\newblock In \emph{International conference on machine learning}, pages 20827--20840. PMLR, 2022.

\bibitem[Sundararajan et~al.(2017)Sundararajan, Taly, and Yan]{sundararajan2017axiomatic}
Mukund Sundararajan, Ankur Taly, and Qiqi Yan.
\newblock Axiomatic attribution for deep networks.
\newblock In \emph{International conference on machine learning}, pages 3319--3328. PMLR, 2017.

\bibitem[Szyc et~al.(2021)Szyc, Walkowiak, and Maciejewski]{szyc2021out}
Kamil Szyc, Tomasz Walkowiak, and Henryk Maciejewski.
\newblock Why out-of-distribution detection in cnns does not like mahalanobis--and what to use instead.
\newblock \emph{arXiv preprint arXiv:2110.07043}, 2021.

\bibitem[Tajwar et~al.(2021)Tajwar, Kumar, Xie, and Liang]{tajwar2021no}
Fahim Tajwar, Ananya Kumar, Sang~Michael Xie, and Percy Liang.
\newblock No true state-of-the-art? ood detection methods are inconsistent across datasets.
\newblock \emph{arXiv preprint arXiv:2109.05554}, 2021.

\bibitem[Tan et~al.(2022)Tan, Hou, Batten, Qiu, Kainz, et~al.]{tan2022detecting}
Jeremy Tan, Benjamin Hou, James Batten, Huaqi Qiu, Bernhard Kainz, et~al.
\newblock Detecting outliers with foreign patch interpolation.
\newblock \emph{Machine Learning for Biomedical Imaging}, 1\penalty0 (April 2022 issue):\penalty0 1--27, 2022.

\bibitem[{U.S. Food and Drug Administration}(2024)]{fda2025guidance}
{U.S. Food and Drug Administration}.
\newblock Marketing submission recommendations for a predetermined change control plan for artificial intelligence/machine learning-enabled device software functions, 2024.
\newblock Accessed: 2025-07-27.

\bibitem[Valiant(1984)]{valiant_theory_1984}
L.~G. Valiant.
\newblock A theory of the learnable‖ {Communications} of the {ACM}, 27 (11): 1134-1142.
\newblock \emph{Google Scholar Google Scholar Digital Library Digital Library}, 1984.

\bibitem[Wang et~al.(2022)Wang, Li, Feng, and Zhang]{wang2022vim}
Haoqi Wang, Zhizhong Li, Litong Feng, and Wayne Zhang.
\newblock Vim: Out-of-distribution with virtual-logit matching.
\newblock In \emph{Proceedings of the IEEE/CVF conference on computer vision and pattern recognition}, pages 4921--4930, 2022.

\bibitem[Wang et~al.(2024)Wang, Lin, Chen, Schmidt, Han, and Zhang]{wang2024sober}
Qizhou Wang, Yong Lin, Yongqiang Chen, Ludwig Schmidt, Bo Han, and Tong Zhang.
\newblock A sober look at the robustness of clips to spurious features.
\newblock \emph{Advances in Neural Information Processing Systems}, 37:\penalty0 122484--122523, 2024.

\bibitem[Yang et~al.(2023)Yang, Zhou, and Liu]{yang2023full}
Jingkang Yang, Kaiyang Zhou, and Ziwei Liu.
\newblock Full-spectrum out-of-distribution detection.
\newblock \emph{International Journal of Computer Vision}, 131\penalty0 (10):\penalty0 2607--2622, 2023.

\bibitem[Yuan et~al.(2023)Yuan, Xia, Dong, Chen, Yao, Qiu, Yan, Yin, Shi, Chen, et~al.]{yuan2023devil}
Mingze Yuan, Yingda Xia, Hexin Dong, Zifan Chen, Jiawen Yao, Mingyan Qiu, Ke Yan, Xiaoli Yin, Yu Shi, Xin Chen, et~al.
\newblock Devil is in the queries: advancing mask transformers for real-world medical image segmentation and out-of-distribution localization.
\newblock In \emph{Proceedings of the IEEE/CVF conference on computer vision and pattern recognition}, pages 23879--23889, 2023.

\bibitem[Zadorozhny et~al.(2022)Zadorozhny, Thoral, Elbers, and Cin{\`a}]{zadorozhny2022out}
Karina Zadorozhny, Patrick Thoral, Paul Elbers, and Giovanni Cin{\`a}.
\newblock Out-of-distribution detection for medical applications: Guidelines for practical evaluation.
\newblock In \emph{Multimodal AI in healthcare: A paradigm shift in health intelligence}, pages 137--153. Springer, 2022.

\bibitem[Zagoruyko and Komodakis(2016)]{zagoruyko2016wide}
Sergey Zagoruyko and Nikos Komodakis.
\newblock Wide residual networks.
\newblock In \emph{Procedings of the British Machine Vision Conference 2016}, pages 87--1. British Machine Vision Association, 2016.

\bibitem[Zhang et~al.(2022)Zhang, Fu, Chen, Du, Li, Wang, Han, Zhang, et~al.]{zhang2022out}
Jinsong Zhang, Qiang Fu, Xu Chen, Lun Du, Zelin Li, Gang Wang, Shi Han, Dongmei Zhang, et~al.
\newblock Out-of-distribution detection based on in-distribution data patterns memorization with modern hopfield energy.
\newblock In \emph{The Eleventh International Conference on Learning Representations}, 2022.

\bibitem[Zhang et~al.(2023)Zhang, Yang, Wang, Wang, Lin, Zhang, Sun, Du, Li, Liu, et~al.]{zhang2023openood}
Jingyang Zhang, Jingkang Yang, Pengyun Wang, Haoqi Wang, Yueqian Lin, Haoran Zhang, Yiyou Sun, Xuefeng Du, Yixuan Li, Ziwei Liu, et~al.
\newblock Openood v1. 5: Enhanced benchmark for out-of-distribution detection.
\newblock In \emph{NeurIPS 2023 Workshop on Distribution Shifts: New Frontiers with Foundation Models}, 2023.

\bibitem[Zhou et~al.(2022)Zhou, Liu, Qiao, Xiang, and Loy]{zhou2022domain}
Kaiyang Zhou, Ziwei Liu, Yu Qiao, Tao Xiang, and Chen~Change Loy.
\newblock Domain generalization: A survey.
\newblock \emph{IEEE transactions on pattern analysis and machine intelligence}, 45\penalty0 (4):\penalty0 4396--4415, 2022.

\end{thebibliography}
    \label{sec:bib}
}

\label{sec:after_bib}

\clearpage
\setcounter{page}{1}
\maketitlesupplementary

\section{Computational Resources}

\noindent All computations were performed on a system equipped with NVIDIA RTX A5000 GPUs (24 GB VRAM), an Intel(R) Xeon(R) W-2275 CPU @ 3.30GHz and 256 GB of system RAM. This hardware configuration provided sufficient computational capacity for training and evaluating the models used in this study. Experiments were implemented using PyTorch 2.6.0 with CUDA 12.4 and cuDNN 9.1.0. To ensure reproducibility during evaluation, deterministic GPU execution was enabled by setting \texttt{torch.backends.cudnn.deterministic = True}.

\section{Information on Model Training}
All primary models were optimised using AdamW \cite{loshchilovdecoupled} with an initial learning rate of 1e-4 and weight decay 0.01. Network weights were initialised using He initialisation, appropriate for ReLU-based architectures. Models were trained using the cross-entropy loss for 600 optimisation steps. A multi-step learning rate schedule was applied via PyTorch’s \texttt{MultiStepLR}, with decay milestones at [150,300,450] epochs. Models were trained with batch size 256. 

We employed five repetitions of five-fold cross-validation to ensure robust evaluation across all experiments. Medical imaging datasets often contain repeated images or multiple images from the same patient, which can lead to data leakage. This can be an issue for evaluating generative-based methods such as DDPM-MSE, which have been shown to exhibit memorisation behaviour \cite{baptista_memorization_2025}. To address this, we enforced strict patient-level separation in the CheXpert dataset, ensuring that all images from the same patient were assigned exclusively to the training set. For the ISIC dataset, where generative models were evaluated, we removed all duplicate images prior to training in order to avoid inflating performance due to memorisation. A list of images in the ISIC dataset used for training are given in the Code Appendix.

\begin{figure*}[t] 
    \centering
    \includegraphics[width=\textwidth]{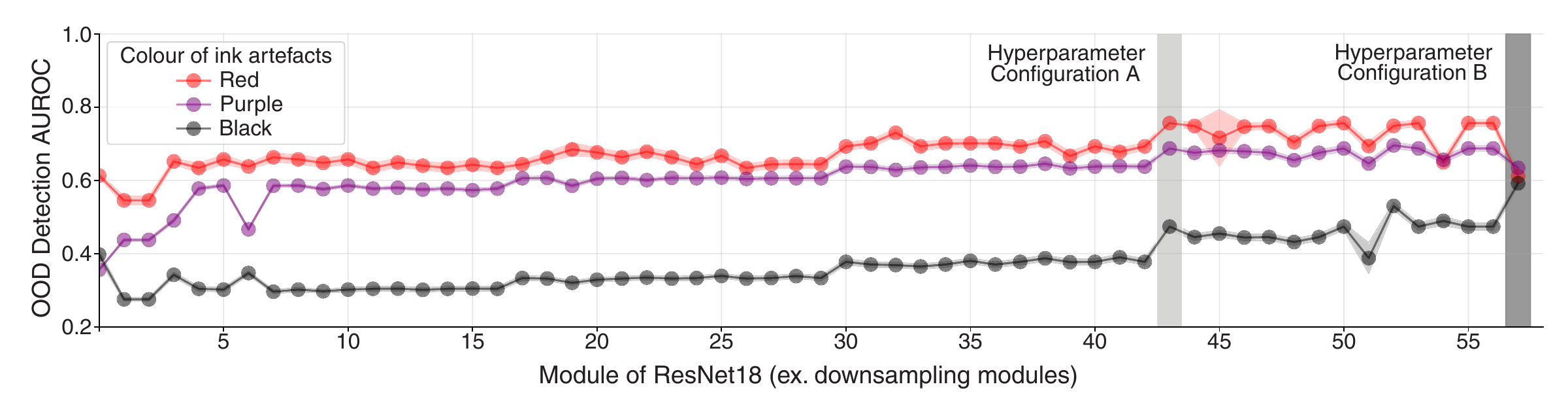} 
    \caption{OOD detection AUROC for Mahalanobis score across ResNet18 modules (excluding downsampling layers) for red, purple and black ink annotations in ISIC. Light (hyperparameter configuration A) and dark grey (hyperparameter configuration B) bands highlight selected modules for comparison. The results show that the observed OOD detection performance varies significantly across layers, demonstrating that conclusions drawn using Mahalanobis score are dependent on the choice of hyperparameter. The Invisible Gorilla Effect would not be observed if only the dark grey hyperparameter is evaluated, highlighting the benefit of evaluating a wide range of hyperparameter settings. Each point represents the mean over 25 seeds, with coloured shaded regions indicating the 95\% confidence interval.  }
    \label{fig:mahal}
\end{figure*}

During training, a series of image augmentations were applied to improve model generalisation. Input images were first resized to 224×224 pixels and then centre-cropped. Spatial augmentations included random rotations of up to 45°, random cropping with up to 25 pixels of padding, horizontal flipping with a probability of 0.5, and random perspective transformations with a distortion scale of 0.2. All images were converted to PyTorch tensors and normalized to float precision. Finally, Channel-wise normalisation was performed using dataset-specific means, which can be accessed in the Code Appendix. These augmentations were applied only during training. During evaluation, only resizing to 224×224 and channel-wise normalisation were applied, with no additional augmentations. Note that no colour-perturbing augmentations were used during training, as this would impact the colour of the model's regions of interest, which was a focus of this study.

For internal ad-hoc methods, the training regime of the primary models were different to increase the OOD-awareness of the model. Bayes by Backprop \cite{blundell2015weight} is a variational inference method for training Bayesian neural networks. A Gaussian prior $\mathcal{N}(0, 1)$ was placed over each weight. Posterior distributions were initialised with mean $\mu = 0$ and scale parameter $\rho = -3$, where the standard deviation was computed as $\sigma = \log(1 + \exp(\rho))$. During each training epoch, we performed five Monte Carlo forward passes and optimised the model using the evidence lower bound (ELBO) \cite{blei2017variational}, which consisted of a cross-entropy loss and a KL divergence loss. CIDER \cite{ming2022exploit} was trained using supervised contrastive learning with a regularisation term based on class prototypes. Prototypes were updated using an exponential moving average (EMA). Two augmented views of each input were passed through the network, and the resulting features were normalised before computing the loss. For CIDER, the total loss was a combination of a discriminative loss and a compactness loss \cite{ming2022exploit}, which we gave equal weighting ($w=1$). For Outlier Exposure \cite{hendrycksdeep}, we applied a KL-divergence loss on the OOD samples, weighted by 0.5, and combined it with the cross-entropy loss for the main image classification task. For Reject Class \cite{chow2003optimum}, we optimised the model using only cross-entropy loss, assigning all OOD samples to the additional class label. For both Outlier Exposure and Reject Class, we used CIFAR-10 \cite{krizhevsky2009learning} as the auxiliary OOD dataset, following the setup from the original Outlier Exposure paper \cite{hendrycksdeep}. Finally, for Rotation Prediction \cite{hendrycks2019using}, an additional head was added to the primary model, to predict the rotation of a training image from the set $\{$ 0°, 90°, 180°, 270°$\}$. A cross-entropy loss was applied to the rotation prediction task, weighted by 0.25, and combined with the cross-entropy loss for the main image classification task.

For external methods, a model was trained external to the primary model trained for the image classification task. For DeepSVDD \cite{ruff2018deep}, we first trained a WideResNet-based \cite{zagoruyko2016wide} autoencoder to reconstruct the input images using mean squared error for 200 epochs. We then used the WideResNet encoder as the feature extractor for anomaly detection. After pre-training, we initialised the centre of the DeepSVDD hypersphere $c \in \mathbb{R}^{32}$ as the mean of the feature embeddings from the training data, where 32 is the dimensionality of the encoder’s output. To avoid trivial solutions, dimensions of $c$ with absolute value below $1 \times 10^{-5}$ were set a value of $1 \times 10^{-5}$. During the main DeepSVDD training phase, we optimised the encoder to minimise the average squared Euclidean distance between the feature representations of the inputs and the centre c. For training the DDPM \cite{graham2023denoising}, we followed the original setup by defining a linear noise schedule over $T = 1000$ timesteps. The variance schedule $\{\beta_t\}_{t=1}^T$ was linearly spaced between a minimum value of $\beta_{\text{min}} = 10^{-4}$ and a maximum value of $\beta_{\text{max}} = 0.02$. At each training step, noise was added to input images, and the model learned to predict this noise using a U-Net \cite{ronneberger2015u} conditioned on the diffusion timestep. The network was optimised using a mean squared error loss. For Foreign Patch Interpolation \cite{tan2022detecting}, each batch consisted of paired image patches extracted from both original and shuffled input images. The model was trained to distinguish whether a patch originated from the same image or from a different one. Patch centres were sampled using a core percent value of 0.99, meaning that only a narrow margin around the image borders was excluded to avoid placing patches near the edges. Training used a pixel-wise binary cross-entropy loss, where the loss was weighted based on the proportion of positive pixels in each batch \cite{tan2022detecting}. Finally, for RealNVP \cite{kirichenko2020normalizing}, the model consisted of 6 affine coupling layers, each with a separate scale and translation network. These networks are feedforward MLPs that operate on a masked portion of the input to compute transformations for the remaining features. After every two coupling layers, a learnable permutation layer is applied to shuffle the feature dimension \cite{kirichenko2020normalizing}. For training the RealNVP model, we extracted feature representations from a selected layer of a pre-trained primary model and computed per-class mean and precision statistics. We use these statistics to remove correlations between features, a process known as whitening. For each class, a separate RealNVP model was trained on the whitened features using maximum likelihood estimation.

\begin{table*}[t]
\centering
\caption{List of hyperparameters for \emph{feature-based OOD detection} methods. The table presents the hyperparameter configurations evaluated in our study, along with the total number of settings used per primary network architecture for each method.}
\begin{tabular}{|m{2.5cm}|m{7cm}|>{\centering\arraybackslash}m{1.3cm}|>{\centering\arraybackslash}m{1.3cm}|>{\centering\arraybackslash}m{1.3cm}|}
\hline
\textbf{Method} & \textbf{Method Hyperparameter(s)} & \multicolumn{3}{c|}{\textbf{Number of Settings per Model}} \\
\cline{3-5}
& & \textbf{ResNet18} & \textbf{VGG16} & \textbf{ViT-b32} \\
\hline
CoP & 
\shortstack[l]{
$\bullet$ Dimensionality, $D_{\text{CoP}} \in [2,10,20]$
}
& 3 & 3 & 3 \\
\hline
CoRP & 
\shortstack[l]{
$\bullet$ Dimensionality, $D_{\text{CoRP}} \in [2,10,20]$\\
$\bullet$ Gaussian Kernel, $\gamma_{\text{CoRP}} = [0.1]$\\
$\bullet$ Number of random Fourier features, $n_{rff} = [20]$ 
}
& 3 & 3 & 3 \\
\hline
FeatureNorm & 
\shortstack[l]{
$\bullet$ Layer selection
}
& 66 & 42 & 139 \\
\hline
GRAM & 
\shortstack[l]{
$\bullet$ Powers, $p_{GRAM} \in [[1],[10],[2,4,6]]$
}
& 3 & 3 & 3 \\
\hline
KDE (Gaussian) & 
\shortstack[l]{
$\bullet$ Layer selection 
}
& 66 & 42 & 139 \\
\hline
KNN & 
\shortstack[l]{
$\bullet$ Layer selection \\
$\bullet$ Number of neighbours, $k_{n} \in [1,3,5,10,20]$
}
& 330 & 210 & 695 \\
\hline
LOF & 
\shortstack[l]{
$\bullet$ Layer selection \\
$\bullet$ Number of neighbours, $k_{n} \in [1,3,5,10,20]$
}
& 330 & 210 & 695 \\
\hline
Mahalanobis & 
\shortstack[l]{
$\bullet$ Layer selection 
}
& 66 & 42 & 139 \\
\hline
MBM & 
\shortstack[l]{
$\bullet$ Bracket, $B \in [1,2,3,4]$
}
& 4 & 4 & 4 \\
\hline
NAN & 
\shortstack[l]{
$\bullet$ Layer selection
}
& 66 & 42 & 139 \\
\hline
NAC & 
\shortstack[l]{
$\bullet$ Layer selection, $l_{\text{NAC}} = $ [Avgpool] \\
$\bullet$ Sigmoid alpha, $\alpha_{\text{NAC}} = [3]$ \\
$\bullet$ Number of bins, $n_{\text{bin-NAC}} = [100]$ 
}
& 1 & 1 & 1 \\
\hline
NMD & 
\shortstack[l]{
$\bullet$ Layer selection
}
& 66 & 42 & 139 \\
\hline
NuSA & 
\shortstack[l]{
$\bullet$ Linear layer, $l_{NuSA} = $ [Avgpool]
}
& 1 & 1 & 1 \\
\hline
PCX & 
\shortstack[l]{
$\bullet$ Layer selection
}
& 21 & 16 & 26 \\
\hline
Residual & 
\shortstack[l]{
$\bullet$ Layer selection \\
$\bullet$ Dimension, $D_{\text{residual}} \in [2,10,20]$
}
& 198 & 126 & 417 \\
\hline
TAPUUD & 
\shortstack[l]{
$\bullet$ Number of clusters, $n_{\text{cluster}} \in [[3,5,7]]$ 
}
& 1 & 1 & 1 \\
\hline
XOOD-M & 
\shortstack[l]{
$\bullet$ Layer selection \\
$\bullet$ Covariance matrix scaling, $C \in [0.1,1.0,10^4]$
}
& 198 & 126 & 417 \\
\hline

\end{tabular}

\label{tab:ood_hyperparams_1}
\end{table*}

\begin{table*}[ht!]
\centering
\caption{List of hyperparameters for \emph{internal ad-hoc OOD detection} methods. The table presents the hyperparameter configurations evaluated in our study.}
\begin{tabular}{|m{2.9cm}|m{7cm}|>{\centering\arraybackslash}m{3.6cm}|}
\hline
\textbf{Method} & \textbf{Method Hyperparameter(s)} & \textbf{Number of Settings} \\
\hline
Bayes By Backprop & 
\shortstack[l]{
$\bullet$ Monte Carlo Samples, $n_{MC} = [30]$\\
$\bullet$ Scoring function, $\mathcal{S}_{\text{BNN}} = $ [Mutual Information]
}
& 1 \\
\hline
CIDER & 
\shortstack[l]{
$\bullet$ Distance function, $d_{\text{CIDER}} = $[Cosine distance] \\
$\bullet$ Feature extractor layer, $l_{\text{CIDER}} = $ [Output layer]
}
& 1 \\
\hline
Outlier Exposure & 
\shortstack[l]{
$\bullet$ Scoring Function, $\mathcal{S}_{\text{OE}} = $ [MCP] 
}
& 1 \\
\hline
Reject Class & 

& 1 \\
\hline
Rotation Prediction & 
\shortstack[l]{
$\bullet$  Scoring Function, $\mathcal{S}_{\text{RP}} = $ [Cross Entropy Loss] 
}
& 1 \\
\hline

\end{tabular}

\label{tab:ood_hyperparams_2}
\end{table*}

\section{Method Hyperparameters}
For each method, we explored a wide range of hyperparameter settings. The specific hyperparameter configurations evaluated in our study are detailed in the following tables: feature-based methods (Table \ref{tab:ood_hyperparams_1}), internal ad-hoc methods (Table \ref{tab:ood_hyperparams_2}),  confidence-based methods (Table \ref{tab:ood_hyperparams_3}), and external methods (Table \ref{tab:ood_hyperparams_4}). To highlight the importance of evaluating a broad range of hyperparameters, we report OOD detection AUROC for the Mahalanobis score across all layers of a ResNet18 on the ISIC benchmark with ink annotations (Figure~\ref{fig:mahal}). While red ink annotations consistently yield higher AUROC scores across most layers, our results show that specific hyperparameter choices can reverse this trend - making red ink annotations no more detectable than purple or black. This illustrates that conclusions about OOD performance can be highly sensitive to the choice of hyperparameters. Therefore, a full evaluation of method performance must consider a comprehensive hyperparameter search to ensure robust and fair comparisons across methods and settings. For each OOD artefact type, we selected the hyperparameter setting that achieved the highest mean performance (e.g. AUROC) for that specific artefact, averaged over 25 random seeds. This procedure was carried out independently for each artefact category, ensuring that the reported results reflect the best-performing configuration for the corresponding artefact.


\begin{table*}[h]
\centering
\caption{The table presents the hyperparameter configurations evaluated in our study for \emph{confidence-based methods}, along with the total number of settings used per primary network architecture for each method. The table presents the hyperparameter configurations evaluated in our study, along with the total number of settings used per primary network architecture for each method.}
\begin{tabular}{|m{2.5cm}|m{7cm}|>{\centering\arraybackslash}m{1.3cm}|>{\centering\arraybackslash}m{1.3cm}|>{\centering\arraybackslash}m{1.3cm}|}
\hline
\textbf{Method} & \textbf{Method Hyperparameter(s)} & \multicolumn{3}{c|}{\textbf{Number of Settings per Model}} \\
\cline{3-5}
& & \textbf{ResNet18} & \textbf{VGG16} & \textbf{ViT-b32} \\
\hline
ASH & 
\shortstack[l]{
$\bullet$ ASH function, $g_{\text{ASH}} \in $[ASH-b, ASH-p, ASH-s] \\
$\bullet$ Percentile, $p_{\text{ASH}} \in [0.6,0.7,0.8,0.9]$
}
& 12 & 12 & 12 \\
\hline
Deep Ensemble & 
\shortstack[l]{
$\bullet$ Number of Ensemble members, $n_{\text{DE}} = [5]$ \\
$\bullet$ Scoring function, $S_{\text{DE}} =$ [MCP]
}
& 1 & 1 & 1 \\
\hline
DICE & 
\shortstack[l]{
$\bullet$ Sparsity parameter, $p_{\text{DICE}} \in [0.6,0.7,0.8,0.9]$ \\
$\bullet$ Truncation layer, $l_{\text{DICE}} = $ [Penultimate layer]
}
& 4 & 4 & 4 \\
\hline
GAIA-A &  & 1 & 1 & 1 \\
\hline
GradNorm & 
\shortstack[l]{
$\bullet$ Parameter selection \\
$\bullet$ Summation method, $\ell_n \in [\ell_1,\ell_2]$
}
& 132 & 64 & 304 \\
\hline
GradOrth & 
\shortstack[l]{
$\bullet$ Epsilon threshold, $\epsilon_{\text{GradOrth}} \in [0.1,0.5,0.9]$
}
& 3 & 3 & 3 \\
\hline
MCP &  & 1 & 1 & 1 \\
\hline
MC-Dropout & 
\shortstack[l]{
$\bullet$ Number of samples, $n_{\text{repeats}} = $ [30] \\
$\bullet$ Dropout probability, $p_{\text{dropout}} \in [0.1,0.2,0.3,0.4]$
}
& 4 & 4 & 4 \\
\hline
ODIN & 
\shortstack[l]{
$\bullet$ Temperature Scaling, $T \in [1,10,100]$ \\
$\bullet$ Preprocessing Magnitude, $\epsilon_{\text{ODIN}} \in$\\
$\{10^{0}, 10^{-1}, 10^{-2}, 10^{-3}, 10^{-4}\}$
}
& 15 & 15 & 15 \\
\hline
ReAct & 
\shortstack[l]{
$\bullet$ Percentile, $p_{\text{ReAct}} \in [0.6,0.7,0.8,0.9]$
}
& 4 & 4 & 4 \\
\hline
SHE &  & 1 & 1 & 1 \\
\hline
ViM & 
\shortstack[l]{
$\bullet$ Weighting coefficient, $\alpha \in [0.25,0.5,0.75]$ \\
$\bullet$ Dimensionality, $D_{\text{ViM}} \in [2,5,10,20]$
}
& 12 & 12 & 12 \\
\hline
WeiPer & 
\shortstack[l]{
$\bullet$ Scoring function, $\mathcal{S}_{\text{WeiPer}} \in $ [MCP, KL-div] \\
$\bullet$ Perturbations, $\delta_{\text{WeiPer}} \in [0.1,1.0,10]$ \\
$\bullet$ Normalising factor, $\epsilon_{\text{WeiPer}} \in [0.01,0.2]$ \\
$\bullet$ Contribution of term 1, $\lambda_1 \in [0.0,1.0,2.5]$ \\
$\bullet$ Contribution of term 2, $\lambda_2 \in [0.0,0.1,1.0]$ \\
$\bullet$ Number of bins, $n_{WeiPer} \in [10,50,100]$ \\
$\bullet$ Number of perturbations, $r_{\text{WeiPer}} = [30]$} 
& 324 & 324 & 324 \\
\hline

\end{tabular}

\label{tab:ood_hyperparams_3}
\end{table*}


\begin{table*}[ht!]
\centering
\caption{List of hyperparameters for \emph{external OOD detection} methods. The table presents the hyperparameter configurations evaluated in our study, along with the total number of settings used per primary network architecture for each method.}
\begin{tabular}{|m{2.9cm}|m{7cm}|>{\centering\arraybackslash}m{3.6cm}|}
\hline
\textbf{Method} & \textbf{Method Hyperparameter(s)} & \textbf{Number of Settings} \\
\hline
DeepSVDD & 
& 1 \\
\hline
DDPM-MSE & 
\shortstack[l]{
$\bullet$ Number of steps, $n_{\text{DDPM-step}} = [1000]$ \\
$\bullet$ Number of reconstructions, $n_{\text{recon}} = [100]$ \\
$\bullet$ ID comparator, $\mathcal{D}_{\text{DDPM}} = $[Training Data] 
}
& 1 \\
\hline
DDPM-LPIPS & 
\shortstack[l]{
$\bullet$ Number of steps, $n_{\text{DDPM-step}} = [1000]$ \\
$\bullet$ Number of reconstructions, $n_{\text{recon}} = [100]$ \\
$\bullet$ ID comparator, $\mathcal{D}_{\text{DDPM}} = $[Training Data] \\
$\bullet$ Feature extractor, $\phi = $[Pretrained AlexNet] 
}
& 1 \\
\hline
Foreign Patch Interpolation & 
\shortstack[l]{
$\bullet$ Threshold, $\tau_{\text{FPI}} \in [0.4,0.5,0.6,0.7,0.8,0.9] $ 
}
& 6 \\
\hline
RealNVP & 
\shortstack[l]{
$\bullet$ Layer selection 
}
& \shortstack[l]{
$\bullet$ 66 for ResNet18 \\
$\bullet$ 42 for VGG16 \\
$\bullet$ 139 for Vit-b32 
} \\
\hline

\end{tabular}

\label{tab:ood_hyperparams_4}
\end{table*}


\section{Annotation Summary}
For the ISIC dataset, we used two out-of-distribution benchmarks: colour charts and ink annotations. We manually annotated 8,964 instances of colour charts and 2,358 instances of ink annotations. To enable our analysis, we manually annotated each of these images by colour. For colour charts, we first annotated each image containing a chart where a specific colour was present (Table~\ref{tab:cc_1}). To evaluate OOD detection performance by colour, we also created a subset where each image contained only a single annotated colour (Table~\ref{tab:cc_2}). All images in this subset were used to generate synthetic, colour-swapped counterfactuals, as described in the Main Paper. For our final analysis, we restricted evaluation to colour charts occupying less than 10\% of the image area (Table~\ref{tab:cc_3}), as large charts are easily detected (achieving near-perfect OOD AUROC across all colours) and thus offer limited insight into the effect of colour on detection performance. Similarly, for ink annotations, we manually annotated images containing any ink color (Table~\ref{tab:ink_1}), as well as a subset of images containing only a single ink colour (Table~\ref{tab:ink_2}). Finally, a summary for the annotations for the MVTec-AD benchmarks is given in (Table~\ref{tab:mvtec_1}).

\begin{table}[H]
\centering
\caption{Number of images containing a colour chart in ISIC where the specified colour is present. For images with multiple colours, each colour is counted independently (i.e. counts are not mutually exclusive).}
\begin{tabular}{|l|c|}
\hline
\textbf{Colour of chart} & \textbf{Number of Images} \\
\hline
Blue & 4\,203 \\
Yellow & 2\,502 \\
Green & 2\,315 \\
Orange & 2\,193 \\
Red & 1\,462 \\
Grey / white & 380 \\
Black & 47 \\
\hline
\end{tabular}

\label{tab:cc_1}
\end{table}

\begin{table}[H]
\centering
\caption{Number of images containing a colour chart in ISIC with a single, uniquely assigned colour. Each image is associated with exactly one colour category.}
\begin{tabular}{|l|c|}
\hline
\textbf{Colour of chart} & \textbf{Number of Images} \\
\hline
Blue & 2\,334 \\
Yellow & 1\,202 \\
Green & 648 \\
Orange & 571 \\
Red & 805 \\
Grey / white & 365 \\
Black & 42 \\
\hline
\end{tabular}

\label{tab:cc_2}
\end{table}

\begin{table}[H]
\centering
\caption{Number of images containing a colour chart in ISIC with a uniquely assigned colour, where the size of the colour chart is \textbf{less than 10\%} of the image area. Each image is associated with exactly one colour category.}
\begin{tabular}{|l|c|}
\hline
\textbf{Colour of chart} & \textbf{Number of Images} \\
\hline
Blue & 904 \\
Yellow & 419 \\
Green & 185 \\
Orange & 207 \\
Red & 321 \\
Grey / white & 92 \\
Black & 5 \\
\hline
\end{tabular}

\label{tab:cc_3}
\end{table}

\begin{table}[H]
\centering
\caption{Number of images containing an ink annotation in ISIC with a uniquely assigned colour where the specified colour is present. For images with multiple colours, each colour is counted independently (i.e. counts are not mutually exclusive).}
\begin{tabular}{|l|c|}
\hline
\textbf{Colour of Ink} & \textbf{Number of Images} \\
\hline
Purple & 1\,558 \\
Black & 887 \\
Green & 26 \\
Red & 25 \\
\hline
\end{tabular}

\label{tab:ink_1}
\end{table}

\begin{table}[H]
\centering
\caption{Number of images containing an ink annotation in ISIC with a single, uniquely assigned colour. Each image is associated with exactly one colour category.}
\begin{tabular}{|l|c|}
\hline
\textbf{Colour of Ink} & \textbf{Number of Images} \\
\hline
Purple & 1\,481 \\
Black & 839 \\
Green & 22 \\
Red & 22 \\
\hline
\end{tabular}
\label{tab:ink_2}
\end{table}

\begin{table}[H]
\centering
\caption{Number of images containing an ink annotation with a single, uniquely assigned colour in the MVTec benchmarks, separated by a) Metal Nuts and b) Pills.}
\begin{tabular}{|l|c|}
\hline
\textbf{Colour of Ink} & \textbf{Number of Images} \\
\hline
\multicolumn{2}{|c|}{a) Metal Nut} \\
\hline
Black & 8 \\
Blue & 8 \\
\hline
\multicolumn{2}{|c|}{b) Pill} \\
\hline
Red & 12 \\
Yellow & 6 \\
\hline
\end{tabular}
\label{tab:mvtec_1}
\end{table}

\section{Evaluating Statistical Significance}
\begin{figure}[t] 
    \centering
    \includegraphics[width=\linewidth]{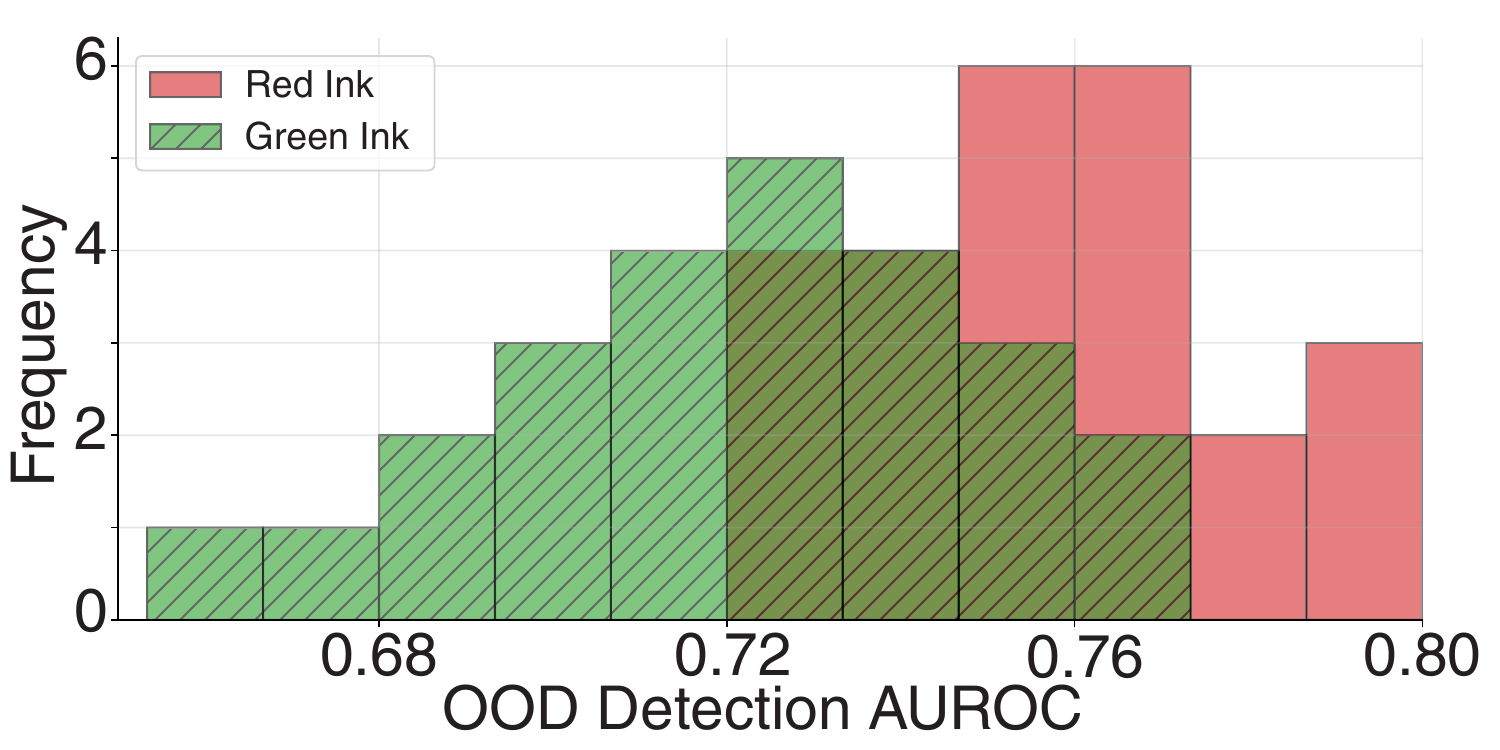} 
    \caption{Distribution of AUROC scores across 25 random seeds for Mahalanobis Score on red (visually similar) and green (visually dissimilar) ink artefacts in the ISIC benchmark. Results are shown for the best-performing feature layer over the OOD dataset. A statistically significant difference was observed ($p = 3.28 \times 10^{-6}$, Wilcoxon signed-rank test). }
    \label{fig:histo}
\end{figure}

To assess the statistical significance of the OOD detection performance difference between similar and dissimilar artefacts, we conducted a two-sided Wilcoxon signed-rank test across 25 random seeds (e.g. 25 models). This non-parametric test was chosen as it does not assume a Gaussian distribution. We compared AUROC scores from red ink annotations (visually similar to skin lesions) and green ink annotations (visually dissimilar), using the best-performing feature layer of the method Mahalanobis Score on the full OOD dataset. Green was selected as the representative dissimilar artefact because Main Paper Figure 1 shows it yielded the highest AUROC among non-similar colours. A histogram of the OOD detection AUROCs per experiment are plotted in Figure \ref{fig:histo}. The resulting p-value was $3.28\times10^{-6}$, indicating a statistically significant difference in detection performance between the two OOD artefact colours.

\section{Quantifying colour similarity with RGB distance}
To quantify colour similarity, we used linear Euclidean RGB distance. While metrics such as CIEDE2000 more closely reflect human colour perception \cite{Luo2001CIEDE2000}, our goal in this study is to analyse how the neural network processes colour similarity within its input representation, which is defined in RGB. Therefore, RGB Euclidean distance was used because it enables a more direct measure of similarity in the space in which the network operates.

\begin{figure}[t] 
    \centering
    \includegraphics[width=\linewidth]{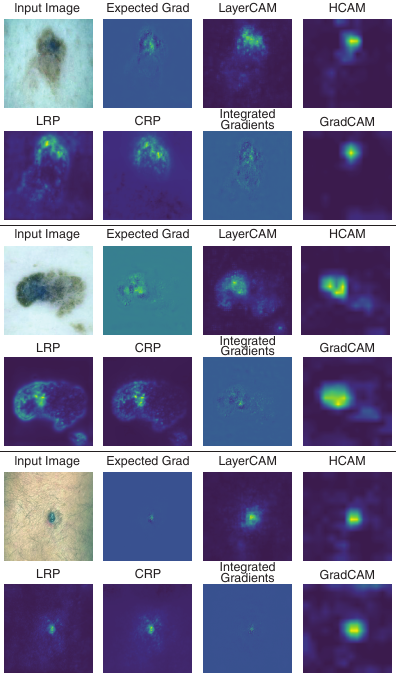} 
    \caption{Explainability analysis of model focus on ISIC lesion regions. Three dermoscopic images from ISIC (top left) and their corresponding saliency maps generated using seven explainable AI methods (Expected Gradients \cite{erion2021improving}, LayerCAM \cite{Jiang2021LayerCAM}, HiResCAM \cite{draelos2020use}, LRP \cite{bach2015pixel}, CRP \cite{achtibat_attribution_2023}, Integrated Gradients \cite{sundararajan2017axiomatic} and GradCAM \cite{selvaraju2017grad}) for a VGG16 primary model trained for classifying between malignant versus benign lesions. cross all methods, the highlighted regions consistently localise on the lesion, providing evidence that the primary models’s region of interest is the lesion. }
    \label{fig:XAI_maps}
\end{figure}

We constructed OOD detection benchmarks to evaluate the Invisible Gorilla Effect. To obtain ROI masks for each dataset, we used the Segment Anything Model (SAM) \cite{kirillov_segment_2023}. We define the ROI as the part of the training image where the task-relevant visual features are expected to reside. For ISIC, we defined the ROI as the lesion area, as this is where features for malignant-benign classification are expected to reside. As classifiers can sometimes use contextual shortcuts (Clever Hans), we verified the model focuses on the lesion region using saliency maps from seven explainable AI methods: Expected Gradients \cite{erion2021improving}, LayerCAM \cite{Jiang2021LayerCAM}, HiResCAM \cite{draelos2020use}, LRP \cite{bach2015pixel}, CRP \cite{achtibat_attribution_2023}, Integrated Gradients \cite{sundararajan2017axiomatic} and GradCAM \cite{selvaraju2017grad} (Figure \ref{fig:XAI_maps}). For MVTec-AD, we defined the ROI as the entire industrial tool (e.g. metal nut).

\begin{figure}[t] 
    \centering
    \includegraphics[width=\linewidth]{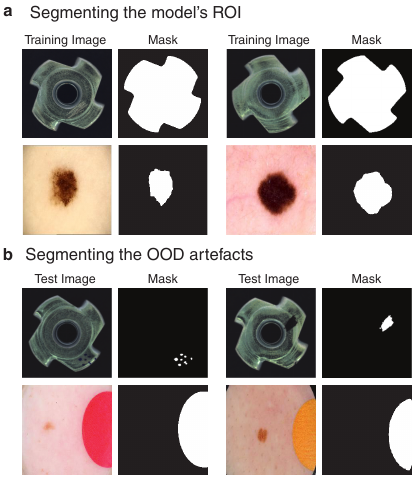} 
    \caption{Examples of segmentation masks used to calculate the mean RGB of regions of interest (ROIs) for in-distribution data and OOD artefacts. A human annotator iteratively prompted the Segment Anything Model (SAM) until the produced mask closely matched the target region. Panel (a) shows segmentations of the model’s ROI and panel (b) shows segmentations the OOD artefacts for the ISIC and MVTec benchmarks. }
    \label{fig:SAM}
\end{figure}

A human annotator iteratively provided prompts to SAM until the generated segmentation closely matched the target object (example segmentations shown in Figure \ref{fig:SAM}). Utilising these segmentation masks, the mean RGB values of the model's ROI in the data were calculated over the training (ID) dataset (summarised in Table~\ref{tab:ROI_colour}). Note that for ISIC, a sample of 100 images was used to calculate the mean RGB, list of images used provided upon paper acceptance.


\begin{table}[H]
\centering
\caption{Mean RGB for the model's ROI across the training/ID dataset (or a sample of 100 images for ISIC), using segmentation masks created with SAM using human annotator-guided prompts.}
\begin{tabular}{|l|l| l | c |}
\hline
\textbf{Dataset} & \textbf{Model ROI} & \textbf{Mask} &  \textbf{Mean RGB}\\
\hline
ISIC & Skin Lesion & SAM & (176, 116, 77) \\
MVTec-AD & Metal Nut & SAM & (77, 86, 83) \\

\hline
\end{tabular}

\label{tab:ROI_colour}
\end{table}

This process was then repeated to segment the naturally occurring OOD artefacts in these datasets. The segmentation masks were either created using SAM with a human annotator, or using provided ground truth masks (for MVTec covariate OOD artefacts). The mean RGB for each artefact was calculated for the artefacts in ISIC (Table~\ref{tbl:cc_colour}) and MVTec (Table~\ref{tbl:Nut_colour}). For each dataset, we then calculated the Euclidean RGB distance between the artefact colour and the model’s ROI colour and applied a dataset-specific threshold to categorise artefacts as similar (distance below threshold) or dissimilar (distance above threshold). This thresholding serves as an operational definition of similarity, as our goal was not to determine an absolute  boundary but to analyse how OOD detection performance varies with colour distance. This allows us to examine the counterintuitive phenomenon where OOD detection may improve as artefact colours become more similar to those found in the training data.

\begin{table}[H]
\centering
\caption{Mean RGB values of the colour-chart artefact colours used to make the OOD benchmark for ISIC. Segmentation masks were created using SAM with human annotator-guided prompts. The Euclidean distance between the mean artefact and model's ROI is given ($\ell_2$), along with the categorisation of similar and dissimilar colours under a threshold of $\ell_2=90$. }
\begin{tabular}{|l| l | c| c | c |}
\hline
\textbf{Colour} & \textbf{Mask} & \textbf{Mean RGB} & \textbf{$\ell_2$ dist} & \textbf{Label} \\
\hline
Red & SAM & (222, 52, 57) & 81.3 & Sim. \\
Orange & SAM & (207, 123, 48) & 43.0 & Sim. \\
Yellow & SAM & (207, 191, 48) & 86.2 & Sim. \\
\hline
Blue & SAM & (65, 52, 57) & 129.7 & Diss. \\
Green & SAM & (53, 152, 69) & 128.4 & Diss. \\
White & SAM & (144, 158, 162) & 100.1 & Diss.  \\
Black & SAM & (66, 61, 60) & 124.2 & Diss. \\
\hline
\end{tabular}

\label{tbl:cc_colour}
\end{table}

\begin{table}[H]
\centering
\caption{Mean RGB values of the ink artefact colours used to make the OOD benchmark for ISIC. Segmentation masks were created using the ground truth (GT) masks provided by MVTec-AD. The Euclidean distance between the mean artefact and model's ROI is given ($\ell_2$), along with the categorisation of similar and dissimilar colours under a threshold of $\ell_2=42$.}
\begin{tabular}{|l| l | c| c | c |}
\hline
\textbf{Colour} & \textbf{Mask} & \textbf{Mean RGB} & \textbf{$\ell_2$ dist} & \textbf{Label} \\
\hline
Blue & GT & (57, 59, 59) & 40.9 & Sim. \\
\hline
Black & GT & (49, 53, 68) & 45.1 & Diss. \\
\hline
\end{tabular}

\label{tbl:Nut_colour}
\end{table}

\section{On the Trade-Off Between OOD Generalisation and OOD Detection}
A debated question in OOD detection is which inputs a model should be expected to generalise to and which should instead be rejected by an OOD detector. Some works argue that models ought to generalise to all covariate-shifted inputs, and OOD detection should exclusively be used to detect semantic shifts \cite{yang2023full,zhou2022domain}. Within this framing, robustness to every form of covariate shift is the desired goal, and evaluating OOD detection methods on covariate-shifted data is considered misaligned with this objective.


We argue that this framing is insufficient for real-world deployment, particularly in high-risk domains such as the focus of this work. In real-world scenarios, covariate shifts can produce inputs that the primary model should \textit{not} generalise to. For example, covariate shifts can produce inputs for which the main classification task becomes ill-posed, even when the label space itself has not changed. Such shifts can hide, distort, or remove the visual cues needed for a reliable clinical decision. As a result, the model may produce high-confidence predictions even though the input no longer contains enough usable information to justify them. This applies even when the model happens to output the correct label: correctness by coincidence does not imply that the prediction is trustworthy, and such cases remain important to detect \cite{anthony_evaluating_2025}. In these cases, enforcing generalisation is undesirable: the desirable behaviour is to recognise that the input falls outside the model’s scope \cite{guerin_out--distribution_2023} and to flag it as OOD.
In such cases, the optimal behaviour is not to encourage the model to generalise to this covariate-shifted data, but instead to label these inputs as OOD and discard the predictions as unreliable. This is why many prior OOD detection works treat covariate shifts as OOD \cite{averly2023unified, zhang2023openood}.  We note that covariate shifts span a continuum from benign corruptions to task-invalidating artefacts, and we recognise that the boundary between robustness and covariate OOD is not universally defined. Therefore, in Section 2.1 in the main paper, we explicitly operationalise the covariate OOD settings to remove ambiguity about which artefacts we aim to detect.

For these reasons, we evaluate OOD detection on covariate-shift benchmarks that produce inputs outside the model’s reliable operating conditions. In our experiments, these naturally occurring artefacts led to a substantial degradation in classification accuracy (main paper, Fig. 4), confirming that they meaningfully impact task performance. In medical imaging, collecting training data that covers the full range of real-world artefacts is often impractical due to privacy and acquisition constraints. Consequently, detecting previously unseen covariate shifts that compromise the reliability of model predictions is essential for safe deployment.



\section{Supplementary Experimental Results for Section 4.2}
In addition to the main results, we conducted supplementary experiments using VGG16 and ViT-B/32 architectures to assess the generality of our findings across different primary model architectures. We report results using three metrics: OOD detection AUROC, OOD detection AURC (Area Under the Risk–Coverage Curve), and FPR@TPR80. These metrics were chosen because they are unaffected by class imbalance between in-distribution and OOD samples, enabling fair comparison of OOD detection performance across artefact colours with differing numbers of images. The metric FPR@TPR80 places a high threshold (many diagnoses discarded) on the scoring function whereas the other metrics (AUROC and AURC) are threshold-independent, highlighting that threshold adjustment does not eliminate the bias. We report the post-hoc method results for the ISIC benchmarks with a ResNet primary model, with metrics AUROC (Table 2 main paper), AURC (Table~\ref{Tbl:AURC}) and FPR@TRP80 (Table~\ref{Tbl:FPR}). Post-hoc method results for the ISIC benchmarks are shown for both a ViT-B/32 primary model (Table~\ref{Tbl:ISIC_vit}) and a VGG16 primary model (Table~\ref{Tbl:ISIC_vgg}). In addition, post-hoc method results for MVTec benchmarks are shown for a ResNet18 primary model (Table~\ref{Tbl:mvtec_resnet}), ViT-B/32 primary model (Table~\ref{Tbl:mvtec_vit}) and VGG16 primary model (Table~\ref{Tbl:mvtec_vgg}).

\section{Supplementary Experimental Results for Section 4.3}
In addition to the main results, we conducted supplementary experiments for the mitigation strategies across different primary model architectures and mitigation method hyperparameters.

\textbf{Colour jitter augmentation:} We report the OOD detection AUROC results for all the post-hoc methods for a ResNet18 primary model (Table \ref{Tbl:cj_resnet}). In addition, we repeated the analysis for ViT-B/32 primary model using the same colour jitter augmentation, with results given in table \ref{Tbl:cj_vit}.

\textbf{Subspace Projection:} We first evaluate the inference latency for a single image across three feature-based OOD detection methods, both with and without subspace projection, on the ISIC Ink benchmark (Table \ref{tab:inf_lat}). The results show that incorporating the subspace projection introduces only a negligible increase in inference time. In contrast, the latency remains substantially lower than that of external generative approaches such as DDPM-MSE.

We then evaluated how setting the number of PC's in the nuisance subspace (k) impacted the OOD detection performance of method Mahalanobis (see Sec 3.4). We evaluated k$\in [2,5,10,20]$, with results plotted in Table~\ref{tab:k_hyper}. The results show that $k=5$ resulted in the highest performing OOD detection performance for both similar and dissimilar artefacts.

Finally, we applied the subspace projection method on a number of feature-based methods for a ResNet18 primary model (Table~\ref{tab:proj_res}). Some feature-based methods, such as NuSA and PCX, would not be impacted by projecting out the nuisance subspace due to the design of the method. Our results show that projected features can reduce the Invisible Gorilla Effect across several feature-based methods (reducing the gap in OOD detection performance between similar and dissimilar artefacts), but not for all methods (e.g. XOOD-M).

\begin{table*}[h]
\centering
\small
\caption{OOD detection \textbf{AURC} (\%) results for the ISIC benchmark, using a \textbf{ResNet18} primary model for internal methods. Detection was evaluated on artefacts that are either visually similar to skin lesions (red for ink artefacts; red, orange, yellow for colour charts) or visually dissimilar (green, purple, black for ink artefacts; green, blue, black, grey for colour charts). Methods are grouped into two groups: feature-based and confidence-based. Each entry shows the best-performing hyperparameter setting, reported as the mean AURC over 25 seeds, with 95\% confidence intervals in brackets. For AURC, lower values indicate better performance.}
\begin{tabular}{lc >{\columncolor[gray]{0.93}}c c >{\columncolor[gray]{0.93}}c}
\toprule
\multirow{3}{*}{\textbf{OOD Method}} & \multicolumn{2}{c}{\textbf{Ink Artefacts}} & \multicolumn{2}{c}{\textbf{Colour Chart Artefacts}} \\
& \multicolumn{2}{c}{\textbf{AURC ($\downarrow$)}} & \multicolumn{2}{c}{\textbf{AURC ($\downarrow$)}} \\ 
\cmidrule(lr){2-3} \cmidrule(lr){4-5}
& Similar & Dissimilar & Similar & Dissimilar \\
\midrule
\multicolumn{2}{l}{\textbf{Feature-based Methods }} & & & \\
CoP & 0.11 (0.01) & 17.34 (0.01) & 2.73 (0.01) & 4.28 (0.01) \\
CoRP & 0.14 (0.03) & 17.39 (0.01) & 3.03 (0.01) & 4.71 (0.01) \\
FeatureNorm & 0.15 (0.01) & 26.01 (0.03) & 1.53 (0.01) & 2.11 (0.02) \\
GRAM & 0.12 (0.01) & 14.91 (0.01) & 8.78 (0.01) & 11.79 (0.01) \\
KDE (Gaussian) & 0.18 (0.01) & 18.74 (0.01) & 2.12 (0.01) & 3.14 (0.01) \\
KNN & 0.06 (0.01) & 14.76 (0.01) & 0.08 (0.01) & 0.13 (0.01) \\
LOF & 0.11 (0.01) & 19.35 (0.01) & 1.70 (0.01) & 2.73 (0.01) \\
Mahalanobis & 0.10 (0.01) & 17.51 (0.01) & 1.23 (0.01) & 1.98 (0.01) \\
MBM & 0.09 (0.01) & 17.01 (0.01) & 1.12 (0.01) & 1.82 (0.01) \\
NAN & 0.16 (0.01) & 26.14 (0.01) & 6.50 (0.02) & 9.39 (0.02) \\
NAC & 0.44 (0.01) & 32.94 (0.01) & 6.52 (0.02) & 8.43 (0.03) \\
NMD & 0.42 (0.01) & 31.17 (0.01) & 7.87 (0.08) & 9.40 (0.08) \\
NuSA & 0.12 (0.01) & 12.85 (0.03) & 10.02 (0.02) & 13.71 (0.03) \\
PCX & 0.11 (0.01) & 19.41 (0.01) & 0.58 (0.01) & 0.86 (0.01) \\
Residual & 0.35 (0.01) & 28.13 (0.01) & 1.78 (0.01) & 2.40 (0.01) \\
TAPUUD & 0.14 (0.01) & 14.36 (0.01) & 5.40 (0.08) & 6.27 (0.08)\\
XOOD-M & 0.13 (0.01) & 17.06 (0.04) & 3.15 (0.01) & 4.94 (0.01) \\
\midrule
\multicolumn{2}{l}{\textbf{Confidence-based Methods }} & & & \\
ASH & 0.17 (0.01) & 14.43 (0.01) & 5.22 (0.01) & 7.27 (0.01) \\
Deep Ensemble & 0.14 (0.01) & 12.99 (0.01) & 7.50 (0.01) & 10.06 (0.01) \\
DICE & 0.15 (0.01) & 14.29 (0.01) & 9.03 (0.01) & 12.09 (0.01)\\
GAIA-A & 0.21 (0.01) & 19.80 (0.01) & 13.10 (0.01) & 17.54 (0.01) \\
GradNorm &  0.14 (0.01) & 14.74 (0.01) & 10.82 (0.08) & 13.36 (0.07) \\
GradOrth &  0.13 (0.01) & 14.04 (0.01) & 8.98 (0.01) & 12.47 (0.01) \\
MCP & 0.17 (0.01) & 15.74 (0.01) & 9.09 (0.01) & 12.27 (0.01) \\
MC-Dropout & 0.31 (0.01) & 21.24 (0.01) & 10.07 (0.01) & 13.26 (0.01) \\
ODIN & 0.11 (0.01) & 13.82 (0.01) & 2.47 (0.01) & 4.09 (0.01) \\
ReAct &  0.21 (0.01) & 20.94 (0.04) & 11.99 (0.02) & 14.91 (0.02) \\
SHE & 0.14 (0.01) & 14.01 (0.01) & 9.31 (0.01) & 12.77 (0.01) \\
ViM & 0.11 (0.01) & 12.92 (0.01) & 2.31 (0.01) & 2.41 (0.01) \\
WeiPer & 0.12 (0.01) & 12.30 (0.01) & 2.32 (0.01) & 2.44 (0.01) \\
\bottomrule
\end{tabular}

\label{Tbl:AURC}
\end{table*}

\begin{table*}[h]
\centering
\small
\caption{OOD detection \textbf{FPR@TPR80} (\%) results for the ISIC benchmark, using a \textbf{ResNet18} primary model for internal methods. Detection was evaluated on artefacts that are either visually similar to skin lesions (red for ink artefacts; red, orange, yellow for colour charts) or visually dissimilar (green, purple, black for ink artefacts; green, blue, black, grey for colour charts). Methods are grouped into two groups: feature-based and confidence-based. Each entry shows the best-performing hyperparameter setting, reported as the mean FPR@TPR80 over 25 seeds, with 95\% confidence intervals in brackets. For FPR@TPR80, lower values indicate better performance.}
\begin{tabular}{lc >{\columncolor[gray]{0.93}}c c >{\columncolor[gray]{0.93}}c}
\toprule
\multirow{3}{*}{\textbf{OOD Method}} & \multicolumn{2}{c}{\textbf{Ink Artefacts}} & \multicolumn{2}{c}{\textbf{Colour Chart Artefacts}} \\
& \multicolumn{2}{c}{\textbf{FPR@TPR80 ($\downarrow$)}} & \multicolumn{2}{c}{\textbf{FPR@TPR80 ($\downarrow$)}} \\ 
\cmidrule(lr){2-3} \cmidrule(lr){4-5}
& Similar & Dissimilar & Similar & Dissimilar \\
\midrule
\multicolumn{2}{l}{\textbf{Feature-based Methods }} & & & \\
CoP & 55.30 (0.01) & 61.85 (0.01) & 22.06 (0.02) & 28.34 (0.03) \\
CoRP & 56.91 (0.03) & 61.91 (0.01) & 26.16 (0.03) & 32.39 (0.03) \\
FeatureNorm & 34.91 (0.02) & 77.85 (0.03) & 60.44 (0.02) & 64.73 (0.04) \\
GRAM & 49.98 (0.01) & 52.08 (0.01) & 73.44 (0.01) & 76.19 (0.01) \\
KDE (Gaussian) & 45.45 (0.01) & 65.60 (0.01) & 41.79 (0.01) & 49.13 (0.01) \\
KNN & 24.05 (0.01) & 56.67 (0.01) & 10.07 (0.01) & 11.44 (0.01) \\
LOF & 36.18 (0.01) & 64.98 (0.01) & 8.70 (0.01) & 12.85 (0.01) \\
Mahalanobis & 54.18 (0.01) & 63.61 (0.01) & 0.54 (0.01) & 1.35 (0.01) \\
MBM & 52.91 (0.01) & 59.46 (0.01) & 0.53 (0.01) & 1.24 (0.01) \\
NAN & 31.64 (0.02) & 84.62 (0.03) & 46.42 (0.07) & 54.48 (0.07) \\
NAC & 87.09 (0.01) & 90.70 (0.01) & 80.65 (0.01) & 81.40 (0.01) \\
NMD & 83.33 (0.02) & 87.69 (0.03) & 33.79 (0.03) & 34.77 (0.03) \\
NuSA & 47.64 (0.05) & 47.95 (0.03) & 82.99 (0.02) & 84.43 (0.02) \\
PCX & 55.09 (0.01) & 59.17 (0.01) & 3.29 (0.01) & 4.53 (0.01) \\
Residual & 84.55 (0.01) & 83.27 (0.01) & 5.73 (0.01) & 6.29 (0.01) \\
TAPUUD & 52.08 (0.01) & 54.41 (0.01) & 3.57 (0.01) & 7.40 (0.01)\\
XOOD-M & 30.00 (0.03) & 54.20 (0.04) & 18.26 (0.03) & 23.92 (0.06) \\
\midrule
\multicolumn{2}{l}{\textbf{Confidence-based Methods }} & & & \\
ASH & 54.55 (0.01) & 55.64 (0.01) & 44.46 (0.05) & 47.51 (0.04) \\
Deep Ensemble & 48.18 (0.01) & 49.19 (0.01) & 72.93 (0.01) & 73.86 (0.01) \\
DICE & 61.46 (0.01) & 61.50 (0.01) & 73.79 (0.01) & 75.69 (0.01)\\
GAIA-A & 55.27 (0.02) & 63.54 (0.02) & 82.29 (0.03) & 83.70 (0.03) \\
GradNorm & 52.91 (0.03) & 53.97 (0.02) & 41.49 (0.02) & 47.94 (0.02) \\
GradOrth & 51.64 (0.01) & 52.36 (0.01) & 72.96 (0.03) & 74.75 (0.02) \\
MCP & 52.00 (0.01) & 52.31 (0.01) & 73.01 (0.03) & 74.76 (0.02) \\
MC-Dropout & 61.82 (0.01) & 68.93 (0.01) & 75.89 (0.01) & 76.98 (0.01) \\
ODIN & 48.00 (0.01) & 52.30 (0.02) & 17.45 (0.01) & 24.59 (0.01) \\
ReAct &  63.45 (0.02) & 70.45 (0.03) & 78.32 (0.05) & 78.47 (0.05) \\
SHE & 51.64 (0.02) & 53.55 (0.01) & 75.54 (0.03) & 77.00 (0.02) \\
ViM & 50.45 (0.01) & 51.55 (0.01) & 62.33 (0.01) & 64.81 (0.01) \\
WeiPer & 50.65 (0.01) & 50.64 (0.01) & 61.64 (0.01) & 62.12 (0.01) \\
\bottomrule
\end{tabular}

\label{Tbl:FPR}
\end{table*}

\begin{table*}[h]
\centering
\small
\caption{OOD detection AUROC (\%) results for the ISIC benchmark, using a \textbf{ViT-B/32} primary model for internal methods. Detection was evaluated on artefacts that are either visually similar to skin lesions (red for ink artefacts; red, orange, yellow for colour charts) or visually dissimilar (green, purple, black for ink artefacts; green, blue, black, grey for colour charts). Methods are grouped into two groups: feature-based and confidence-based. Each entry shows the best-performing hyperparameter setting, reported as the mean AUROC over 25 seeds, with 95\% confidence intervals in brackets.}
\begin{tabular}{lc >{\columncolor[gray]{0.93}}c c >{\columncolor[gray]{0.93}}c}
\toprule
\multirow{2}{*}{\textbf{OOD Method}} & \multicolumn{2}{c}{\textbf{Ink Artefacts}} & \multicolumn{2}{c}{\textbf{Colour Chart Artefacts}} \\
\cmidrule(lr){2-3} \cmidrule(lr){4-5}
& Similar & Dissimilar & Similar & Dissimilar \\
\midrule
\multicolumn{2}{l}{\textbf{Feature-based Methods }} & & & \\
CoP & 77.43 (0.02) & 72.08 (0.01) & 91.09 (0.01) & 87.99 (0.02) \\
CoRP & 77.51 (0.03) & 70.05 (0.01) & 89.35 (0.01) & 85.52 (0.02) \\
FeatureNorm & 68.37 (0.03) & 52.33 (0.03) & 86.06 (0.01) & 76.39 (0.04) \\
GRAM & 63.54 (0.03) & 52.68 (0.01) & 78.90 (0.03) & 77.92 (0.03) \\
KDE (Gaussian) & 73.89 (0.04) &  64.14 (0.03) & 85.72 (0.03) & 84.27 (0.01) \\
KNN & 82.22 (0.01) & 70.33 (0.01) & 91.48 (0.02) & 91.00 (0.01) \\
LOF & 91.29 (0.01) & 76.50 (0.01) & 92.86 (0.02) & 92.71 (0.01) \\
Mahalanobis & 79.49 (0.02) & 69.39 (0.02) & 91.46 (0.02) & 91.18 (0.02) \\
MBM & 81.20 (0.01) & 68.50 (0.01) & 91.58 (0.02) & 91.48 (0.02) \\
NAN & 53.07 (0.03) & 52.02 (0.01) & 55.12 (0.02) & 54.61 (0.02) \\
NAC & 45.86 (0.02) & 45.32 (0.02) & 57.25 (0.02) & 56.13 (0.02) \\
NMD & 57.88 (0.05) & 57.53 (0.02) & 56.68 (0.08) & 56.26 (0.03) \\
NuSA & 69.97 (0.04) & 68.36 (0.02) & 67.14 (0.04) & 67.11 (0.01) \\
PCX & 60.48 (0.04) & 56.79 (0.01) & 55.81 (0.02) & 55.47 (0.02) \\
Residual & 80.72 (0.02) & 73.86 (0.01) & 91.50 (0.01) & 90.67 (0.01) \\
TAPUUD & 57.70 (0.02) & 55.27 (0.02) & 68.64 (0.02) & 67.93 (0.03) \\
XOOD-M & 65.35 (0.04) & 56.71 (0.01) & 86.79 (0.02) & 86.09 (0.04) \\
\midrule
\multicolumn{2}{l}{\textbf{Confidence-based Methods }} & & & \\
ASH & 69.49 (0.03) & 70.15 (0.01) & 61.68 (0.02) & 61.74 (0.02) \\
Deep Ensemble & 75.48 (0.02) & 72.58 (0.01) & 72.02 (0.01) & 70.83 (0.02) \\
DICE & 67.92 (0.01) & 67.71 (0.01) & 60.81 (0.02) & 60.30 (0.02) \\
GAIA-A & 34.78 (0.06) & 31.19 (0.01) & 34.20 (0.02) & 37.18 (0.02) \\
GradNorm & 71.19 (0.03) & 70.70 (0.01) & 64.94 (0.03) & 63.53 (0.04) \\
GradOrth & 68.28 (0.02) & 70.20 (0.01) & 59.82 (0.03) & 62.29 (0.02) \\
MCP & 69.18 (0.02) & 69.08 (0.01) & 58.72 (0.02) & 58.64 (0.02) \\
MC-Dropout & 72.32 (0.02) & 72.14 (0.01) & 60.23 (0.03) & 59.34 (0.02) \\
ODIN & 71.33 (0.03) & 70.35 (0.01) & 61.29 (0.03) & 61.40 (0.02) \\
ReAct & 67.89 (0.04) & 66.16 (0.03) & 58.96 (0.04) & 57.17 (0.05) \\
SHE & 69.98 (0.03) & 69.63 (0.01) & 58.67 (0.03) & 58.87 (0.02) \\
ViM & 70.65 (0.03) & 70.14 (0.01) & 59.91 (0.03) & 60.74 (0.02) \\
WeiPer & 69.98 (0.02) & 66.31 (0.02) & 61.26 (0.02) & 60.57 (0.02) \\
\bottomrule
\end{tabular}
\label{Tbl:ISIC_vit}
\end{table*}

\begin{table*}[ht!]
\centering
\small
\caption{OOD detection AUROC (\%) results for the ISIC benchmark, using a \textbf{VGG16} primary model for internal methods. Detection was evaluated on artefacts that are either visually similar to skin lesions (red for ink artefacts; red, orange, yellow for colour charts) or visually dissimilar (green, purple, black for ink artefacts; green, blue, black, grey for colour charts). Methods are grouped into two groups: feature-based and confidence-based. Each entry shows the best-performing hyperparameter setting, reported as the mean AUROC over 25 seeds, with 95\% confidence intervals in brackets.}
\begin{tabular}{lc >{\columncolor[gray]{0.93}}c c >{\columncolor[gray]{0.93}}c}
\toprule
\multirow{2}{*}{\textbf{OOD Method}} & \multicolumn{2}{c}{\textbf{Ink Artefacts}} & \multicolumn{2}{c}{\textbf{Colour Chart Artefacts}} \\
\cmidrule(lr){2-3} \cmidrule(lr){4-5}
& Similar & Dissimilar & Similar & Dissimilar \\
\midrule
\multicolumn{2}{l}{\textbf{Feature-based Methods }} & & & \\
CoP & 64.07 (0.01) & 57.30 (0.01) & 54.70 (0.02) & 54.38 (0.02) \\
CoRP & 63.63 (0.01) & 56.07 (0.01) & 54.63 (0.02) & 54.16 (0.02) \\
FeatureNorm & 82.48 (0.01) & 56.18 (0.01) & 60.49 (0.02) & 59.88 (0.03) \\
GRAM & 68.25 (0.04) & 55.25 (0.03) & 59.32 (0.02) & 59.27 (0.02) \\
KDE (Gaussian) & 80.71 (0.01) & 55.74 (0.02) & 60.10 (0.02) & 58.37 (0.02) \\
KNN & 83.67 (0.01) & 62.11 (0.01) & 81.32 (0.01) & 74.24 (0.03) \\
LOF & 85.28 (0.01) & 64.28 (0.01) & 82.38 (0.04) & 82.11 (0.02) \\
Mahalanobis & 80.02 (0.01) & 52.41 (0.01) & 74.38 (0.03) & 74.10 (0.02) \\
MBM & 81.33 (0.01) & 52.06 (0.01) & 75.39 (0.02) & 75.01 (0.02) \\
NAN & 77.25 (0.01) & 48.08 (0.01) & 45.54 (0.03) & 44.71 (0.03) \\
NAC & 51.52 (0.02) & 45.23 (0.01) & 55.63 (0.02) & 55.34 (0.02) \\
NMD & 60.13 (0.04) & 58.24 (0.05) & 37.50 (0.02) & 37.29 (0.02) \\
NuSA & 60.27 (0.06) & 58.03 (0.05) & 82.91 (0.02) & 82.82 (0.02) \\
PCX & 78.33 (0.04) & 63.15 (0.05) & 71.23 (0.02) & 70.99 (0.02) \\
Residual & 42.64 (0.01) & 42.75 (0.01) & 60.19 (0.02) & 59.13 (0.02) \\
TAPUUD & 62.47 (0.03) & 53.79 (0.03) & 61.17 (0.01) & 60.69 (0.01) \\
XOOD-M & 80.77 (0.01) & 58.62 (0.01) & 78.94 (0.02) & 76.37 (0.02) \\
\midrule
\multicolumn{2}{l}{\textbf{Confidence-based Methods }} & & & \\
ASH & 65.03 (0.01) &  65.87 (0.01) & 41.90 (0.02) & 41.88 (0.01) \\
Deep Ensemble & 65.72 (0.06) & 65.12 (0.04) & 47.70 (0.01) & 47.75 (0.02) \\
DICE & 72.92 (0.01) & 72.22 (0.01) & 40.94 (0.02) & 40.44 (0.01) \\
GAIA-A & 48.85 (0.01) & 34.86 (0.03) & 42.75 (0.01) & 42.29 (0.02) \\
GradNorm & 62.24 (0.01) & 62.13 (0.01) & 47.83 (0.02) & 46.86 (0.02) \\
GradOrth & 65.76 (0.01) & 65.99 (0.01) & 43.26 (0.02) & 42.35 (0.02) \\
MCP & 60.12 (0.01) & 59.82 (0.01) & 45.45 (0.01) & 45.04 (0.01) \\
MC-Dropout & 61.34 (0.01) & 61.42 (0.01) & 46.17 (0.01) & 46.09 (0.02) \\
ODIN & 67.89 (0.01) & 66.03 (0.01) & 58.94 (0.02) & 57.88 (0.02) \\
ReAct & 72.81 (0.01) & 72.60 (0.01) & 50.95 (0.03) & 50.78 (0.04) \\
SHE & 65.37 (0.01) & 64.64  (0.01) & 42.20 (0.02) & 41.43 (0.02) \\
ViM & 66.43 (0.01) & 66.12 (0.02) & 42.47 (0.02) & 42.60 (0.02) \\
WeiPer & 59.30 (0.01) & 52.34 (0.01) & 53.47 (0.02) & 53.19 (0.02) \\
\bottomrule
\end{tabular}
\label{Tbl:ISIC_vgg}
\end{table*}

\begin{table*}[ht!]
\centering
\small
\caption{OOD detection AUROC (\%) results for the MVTec benchmark, using a \textbf{ResNet18} primary model for internal methods. Detection was evaluated on artefacts that are either visually similar to the model's ROI (red for pill; black for metal nut) or visually dissimilar (yellow for ink artefacts; blue for metal nut). Methods are grouped into two groups: feature-based and confidence-based. Each entry shows the best-performing hyperparameter setting, reported as the mean AUROC over 25 seeds, with 95\% confidence intervals in brackets.}
\begin{tabular}{lc >{\columncolor[gray]{0.93}}c c >{\columncolor[gray]{0.93}}c}
\toprule
\multirow{2}{*}{\textbf{OOD Method}} & \multicolumn{2}{c}{\textbf{Pill}} & \multicolumn{2}{c}{\textbf{Metal Nut}} \\
\cmidrule(lr){2-3} \cmidrule(lr){4-5}
& Similar & Dissimilar & Similar & Dissimilar \\
\midrule
\multicolumn{2}{l}{\textbf{Feature-based Methods }} & & & \\
CoP & 88.27 (0.01) & 81.15 (0.02) & 53.07 (0.01) & 39.20 (0.03) \\
CoRP & 81.92 (0.01) & 71.41 (0.02) & 48.86 (0.01) & 39.09 (0.03) \\
FeatureNorm & 71.67 (0.02) & 56.41 (0.02) & 55.00 (0.01) & 52.69 (0.02) \\
GRAM & 74.76 (0.01) & 69.83 (0.01) & 68.94 (0.01) & 64.70 (0.01) \\
KDE (Gaussian) & 81.35 (0.02) & 73.59 (0.02) & 63.52 (0.01) & 43.98 (0.01) \\
KNN & 93.33 (0.01) & 86.15 (0.01) & 71.02 (0.01) & 36.93 (0.01) \\
LOF & 83.78 (0.01) & 62.18 (0.01) & 69.30 (0.02) & 42.95  (0.02)\\
Mahalanobis & 71.86 (0.01) & 68.72 (0.01) & 69.77 (0.01)  & 58.30 (0.01) \\
MBM & 73.18 (0.01)  & 71.36 (0.01) & 70.32 (0.02) & 58.12 (0.01) \\
NAN & 64.10 (0.02) & 56.92 (0.01) & 54.43 (0.01) & 52.39 (0.03)\\
NAC & 68.27 (0.02) & 67.88 (0.01) & 52.38 (0.01) & 51.93 (0.02) \\
NMD & 85.26 (0.01) & 82.31 (0.01) & 68.75 (0.01) & 68.94 (0.01) \\
NuSA & 80.64 (0.01) & 73.21 (0.01) & 58.41 (0.01) & 44.09 (0.02) \\
PCX & 64.10 (0.02) & 56.41 (0.02) & 51.42 (0.01) & 49.15 (0.01) \\
Residual & 86.09 (0.01) & 68.97 (0.02) & 78.30 (0.03) & 70.91 (0.03) \\
TAPUUD & 53.14 (0.03) & 49.87 (0.04) & 59.27 (0.01) & 58.30 (0.02) \\
XOOD-M & 75.71 (0.02) & 66.67 (0.01) & 67.16 (0.01) & 66.93 (0.01) \\
\midrule
\multicolumn{2}{l}{\textbf{Confidence-based Methods }} & & & \\
ASH & 81.35 (0.02) & 80.64 (0.01) & 57.84 (0.01) & 58.75 (0.02) \\
Deep Ensemble & 79.54 (0.01) & 79.51 (0.01) & 64.72 (0.01) & 62.75 (0.01) \\
DICE & 87.63 (0.01) & 82.56 (0.01) & 63.52 (0.01) & 48.98 (0.02) \\
GAIA-A & 36.47 (0.01) & 32.18 (0.01) & 59.43 (0.02) & 42.61 (0.02) \\
GradNorm & 80.13 (0.01) & 79.07 (0.01) & 60.34 (0.01) & 59.83 (0.01) \\
GradOrth & 86.86 (0.01) & 78.59 (0.01) & 65.45 (0.01) & 53.00 (0.02) \\
MCP & 78.46 (0.02) & 78.33 (0.01) & 58.75 (0.01) & 45.34 (0.01) \\
MC-Dropout & 79.57 (0.02) & 79.04 (0.01) & 60.74 (0.02) & 60.83 (0.02) \\
ODIN & 82.69 (0.01) & 80.77 (0.01) & 58.75 (0.01) & 53.18 (0.02) \\
ReAct & 83.72 (0.01) & 80.00 (0.01) & 59.43 (0.01) & 45.91 (0.03) \\
SHE & 81.09 (0.01) & 80.02 (0.01) & 57.50 (0.01) & 46.02 (0.01) \\
ViM & 83.21 (0.01) & 80.77 (0.01) & 57.27 (0.02) & 43.75 (0.02) \\
WeiPer & 80.77 (0.02) & 80.29 (0.01) & 67.91 (0.01) & 66.02 (0.01) \\

\bottomrule
\end{tabular}
\label{Tbl:mvtec_resnet}
\end{table*}

\begin{table*}[ht!]
\centering
\small
\caption{OOD detection AUROC (\%) results for the MVTec benchmark, using a \textbf{ViT-B/32} primary model for internal methods. Detection was evaluated on artefacts that are either visually similar to the model's ROI (red for pill; black for metal nut) or visually dissimilar (yellow for ink artefacts; blue for metal nut). Methods are grouped into two groups: feature-based and confidence-based. Each entry shows the best-performing hyperparameter setting, reported as the mean AUROC over 25 seeds, with 95\% confidence intervals in brackets.}
\begin{tabular}{lc >{\columncolor[gray]{0.93}}c c >{\columncolor[gray]{0.93}}c}
\toprule
\multirow{2}{*}{\textbf{OOD Method}} & \multicolumn{2}{c}{\textbf{Pill}} & \multicolumn{2}{c}{\textbf{Metal Nut}} \\
\cmidrule(lr){2-3} \cmidrule(lr){4-5}
& Similar & Dissimilar & Similar & Dissimilar \\
\midrule
\multicolumn{2}{l}{\textbf{Feature-based Methods }} & & & \\
CoP & 72.76 (0.01) & 66.92 (0.01) & 78.79 (0.02) & 42.33 (0.02) \\
CoRP & 69.87 (0.01) & 64.38 (0.01) & 75.19 (0.03) & 41.86 (0.02) \\
FeatureNorm & 66.03 (0.02) & 62.49 (0.02) & 85.23 (0.02) & 68.75 (0.01) \\
GRAM & 57.55 (0.03) & 54.83 (0.02) & 78.23 (0.02) & 52.54 (0.02) \\
KDE (Gaussian) & 66.35 (0.01) & 48.85 (0.01) & 82.39 (0.02) & 70.80 (0.02) \\
KNN & 77.95 (0.02) & 71.28 (0.02) & 83.18 (0.02) & 78.75 (0.01) \\
LOF & 63.01 (0.01) & 57.69 (0.01) & 81.59 (0.01) & 79.66 (0.01) \\
Mahalanobis & 70.28 (0.02) & 65.22 (0.02) & 77.70 (0.01) & 67.84 (0.02) \\
MBM & 70.09 (0.03) & 66.27 (0.02) & 78.21 (0.01) & 66.90 (0.01) \\
NAN & 60.90 (0.01) & 58.85 (0.02) & 68.64 (0.01) & 57.39 (0.02) \\
NAC & 48.29 (0.02) & 43.21 (0.02) & 54.54 (0.02) & 52.13 (0.03) \\
NMD & 67.37 (0.01) & 64.87 (0.02) & 70.57 (0.03) & 64.43 (0.02) \\
NuSA & 70.64 (0.01) & 59.10 (0.02) & 58.07 (0.01) & 40.23 (0.01) \\
PCX & 65.87 (0.02) & 63.15 (0.02) & 64.02 (0.03) & 57.20 (0.03) \\
Residual & 66.03 (0.01) & 63.72 (0.01) & 74.77 (0.04) & 70.11 (0.02) \\
TAPUUD & 68.90 (0.02) & 65.64 (0.02) & 65.11 (0.02) & 53.64 (0.01) \\
XOOD-M & 70.08 (0.03) & 68.73 (0.02) & 79.97 (0.03) & 53.27 (0.02) \\
\midrule
\multicolumn{2}{l}{\textbf{Confidence-based Methods }} & & & \\
ASH & 61.25 (0.02) & 60.51 (0.01) & 62.84 (0.01) & 61.70 (0.01) \\
Deep Ensemble & 66.43 (0.02) & 64.75 (0.02) & 70.14 (0.04) & 63.48 (0.02) \\
DICE & 58.21 (0.02) & 51.54 (0.02) & 71.82 (0.04) & 66.93 (0.05) \\
GAIA-A & 52.05 (0.02) & 44.29 (0.02) & 37.16 (0.02) & 35.80 (0.01) \\
GradNorm & 65.94 (0.03) & 62.46 (0.04) & 71.75 (0.03) & 69.39 (0.02) \\
GradOrth & 59.46 (0.03) & 59.02 (0.02) & 69.03 (0.02) & 67.99 (0.03) \\
MCP & 63.72 (0.01) & 55.26 (0.01) & 66.14 (0.01) & 62.16 (0.02) \\
MC-Dropout & 66.27 (0.02) & 54.96 (0.02) & 68.56 (0.02) & 66.47 (0.02) \\
ODIN & 63.97 (0.01) & 63.91 (0.01) & 71.25 (0.04) & 62.16 (0.02) \\
ReAct & 54.36 (0.02) & 34.42 (0.02) & 63.18 (0.01) & 56.59 (0.01) \\
SHE & 70.26 (0.02) & 59.04 (0.02) & 58.30 (0.01) & 44.09 (0.02) \\
ViM & 55.45 (0.02) & 37.44 (0.02) & 63.86 (0.01) & 57.61 (0.01) \\
WeiPer & 64.04 (0.02) & 62.18 (0.02) & 66.02 (0.01) & 62.27 (0.02) \\
\bottomrule
\end{tabular}

\label{Tbl:mvtec_vit}
\end{table*}

\begin{table*}[ht!]
\centering
\small
\caption{OOD detection AUROC (\%) results for the MVTec benchmark, using a \textbf{VGG16} primary model for internal methods. Detection was evaluated on artefacts that are either visually similar to the model's ROI (red for pill; black for metal nut) or visually dissimilar (yellow for ink artefacts; blue for metal nut). Methods are grouped into two groups: feature-based and confidence-based. Each entry shows the best-performing hyperparameter setting, reported as the mean AUROC over 25 seeds, with 95\% confidence intervals in brackets.}
\begin{tabular}{lc >{\columncolor[gray]{0.93}}c c >{\columncolor[gray]{0.93}}c}
\toprule
\multirow{2}{*}{\textbf{OOD Method}} & \multicolumn{2}{c}{\textbf{Pill}} & \multicolumn{2}{c}{\textbf{Metal Nut}} \\
\cmidrule(lr){2-3} \cmidrule(lr){4-5}
& Similar & Dissimilar & Similar & Dissimilar \\
\midrule
\multicolumn{2}{l}{\textbf{Feature-based Methods }} & & & \\
CoP & 76.68 (0.04) & 69.07 (0.03) & 68.18 (0.02) & 53.52 (0.02) \\
CoRP & 69.42 (0.02) & 65.77 (0.02) & 63.63 (0.02) & 50.90 (0.04) \\
FeatureNorm & 52.28 (0.02) & 51.15 (0.01) & 57.95 (0.02) & 45.45 (0.04) \\
GRAM & 64.90 (0.03) & 60.19 (0.03) & 72.46 (0.04) & 65.32 (0.03) \\
KDE (Gaussian) & 63.85 (0.03) & 54.74 (0.02) & 69.89 (0.04) & 58.75 (0.01) \\
KNN & 91.28 (0.03) & 82.82 (0.02) & 80.57 (0.03) & 74.89 (0.03) \\
LOF & 71.79 (0.01) & 68.72 (0.02) & 70.00 (0.02) & 63.52 (0.02) \\
Mahalanobis & 71.15 (0.01) & 69.78 (0.02) & 60.80 (0.01) & 54.89 (0.04) \\
MBM & 72.03 (0.01) & 69.58 (0.02) & 61.28 (0.02) & 55.76 (0.02) \\
NAN & 51.77 (0.06) & 50.26 (0.02) & 53.98 (0.04) & 46.48 (0.03) \\
NAC & 52.10 (0.03) & 50.08 (0.02) & 43.53 (0.04) & 41.42 (0.03) \\
NMD & 60.51 (0.02) & 57.95 (0.01) & 82.16 (0.03) & 74.09 (0.02) \\
NuSA & 55.64 (0.02) & 49.10 (0.04) & 61.16 (0.02) & 58.92 (0.03) \\
PCX & 68.56 (0.02) & 66.29 (0.03) & 74.97 (0.02) & 64.52 (0.02) \\
Residual & 79.87 (0.01) & 76.09 (0.04) & 85.11 (0.02) & 77.95 (0.02) \\
TAPUUD & 59.36 (0.02) & 58.46 (0.02) & 52.36 (0.02) & 48.30 (0.02) \\
XOOD-M & 74.08 (0.03) & 68.84 (0.03) & 74.20 (0.04) & 54.89 (0.2) \\
\midrule
\multicolumn{2}{l}{\textbf{Confidence-based Methods }} & & & \\
ASH & 60.45 (0.02) & 60.13 (0.02) & 70.80 (0.02) & 66.36 (0.03) \\
Deep Ensemble & 60.32 (0.02) & 60.62 (0.03) & 76.77 (0.03) & 75.34 (0.03) \\
DICE & 57.05 (0.03) & 53.72 (0.02) & 75.91 (0.02) & 69.66 (0.01) \\
GAIA-A & 48.33 (0.01) & 41.73 (0.02) & 55.57 (0.02) & 45.57 (0.02) \\
GradNorm & 61.94 (0.03) & 58.42 (0.02) & 63.14 (0.02) & 59.98 (0.02) \\
GradOrth & 64.22 (0.02) & 60.18 (0.03) & 59.09 (0.03) & 51.70 (0.03) \\
MCP & 57.95 (0.02) & 53.59 (0.02) & 73.98 (0.04) & 71.07 (0.03) \\
MC-Dropout & 58.32 (0.03) & 55.31 (0.02) & 72.89 (0.02) & 71.30 (0.02) \\
ODIN & 58.46 (0.01) & 53.78 (0.02) & 78.07 (0.01) & 73.98 (0.03) \\
ReAct & 54.87 (0.02) & 51.79 (0.02) & 75.23 (0.04) & 66.48 (0.01) \\
SHE & 56.54 (0.02) & 54.10 (0.02) & 67.05 (0.02) & 52.84 (0.02) \\
ViM & 85.90 (0.01) & 76.18 (0.03) & 75.91 (0.03) & 69.89 (0.01) \\
WeiPer & 58.85 (0.01) & 53.59 (0.02) & 75.34 (0.03) & 74.32 (0.04) \\
\bottomrule
\end{tabular}

\label{Tbl:mvtec_vgg}
\end{table*}

\begin{table*}[h]
\centering
\small
\caption{OOD AUROC (\%) on the ISIC ink benchmark using 10 \textbf{ResNet18} primary models, trained with either light or heavy \textbf{colour jitter} augmentations. Brackets show the OOD AUROC difference compared to the primary model's trained without colour jitter augmentations (Table 2, main paper).}
\begin{tabular}{lc >{\columncolor[gray]{0.93}}c c >{\columncolor[gray]{0.93}}c}
\toprule
\multirow{2}{*}{\textbf{OOD Method}} & \multicolumn{2}{c}{\textbf{Light Augmentation}} & \multicolumn{2}{c}{\textbf{Heavy Augmentation}} \\
\cmidrule(lr){2-3} \cmidrule(lr){4-5}
& Similar & Dissimilar & Similar & Dissimilar \\
\midrule
\multicolumn{5}{l}{\textbf{Feature-based Methods}} \\
CoP & 77.04 \textcolor{darkgreen}{(+5.30)} & 70.34 \textcolor{darkgreen}{(+4.58)} & 79.58 \textcolor{darkgreen}{(+7.84)} & 69.81 \textcolor{darkgreen}{(+4.05)} \\
CoRP & 74.60 \textcolor{darkgreen}{(+3.44)} & 68.72 \textcolor{darkgreen}{(+3.74)} & 74.33 \textcolor{darkgreen}{(+3.17)} & 67.00 \textcolor{darkgreen}{(+2.02)} \\
FeatureNorm & 75.92 \textcolor{darkgreen}{(+0.80)} & 64.24 \textcolor{darkgreen}{(+11.33)} & 75.66 \textcolor{darkgreen}{(+0.54)} & 68.09 \textcolor{darkgreen}{(+15.18)} \\
GRAM & 67.25 \textcolor{darkred}{(-13.07)} & 59.24 \textcolor{darkred}{(-13.47)} & 66.83 \textcolor{darkred}{(-13.49)} & 61.21 \textcolor{darkred}{(-11.50)} \\
KDE (Gaussian) & 89.70 \textcolor{darkgreen}{(+4.15)} & 75.67 \textcolor{darkgreen}{(+7.51)} & 87.43 \textcolor{darkgreen}{(+1.88)} & 76.89 \textcolor{darkgreen}{(+8.73)} \\
KNN & 90.13 \textcolor{darkgreen}{(+4.45)} & 77.26 \textcolor{darkgreen}{(+7.16)} & 87.88 \textcolor{darkgreen}{(+2.20)} & 79.23 \textcolor{darkgreen}{(+9.13)} \\
LOF & 84.12 \textcolor{darkgreen}{(+1.77)} & 74.52 \textcolor{darkgreen}{(+11.62)} & 88.84 \textcolor{darkgreen}{(+6.49)} & 74.73 \textcolor{darkgreen}{(+11.83)} \\
Mahalanobis & 85.55 \textcolor{darkgreen}{(+8.57)} & 70.92 \textcolor{darkgreen}{(+7.28)} & 87.32 \textcolor{darkgreen}{(+10.34)} & 75.16 \textcolor{darkgreen}{(+11.52)} \\
MBM & 85.52 \textcolor{darkgreen}{(+8.12)} & 71.39 \textcolor{darkgreen}{(+7.68)} & 87.17 \textcolor{darkgreen}{(+9.77)} & 73.88 \textcolor{darkgreen}{(+10.17)} \\
NAN & 76.03 \textcolor{darkgreen}{(+0.45)} & 58.88 \textcolor{darkgreen}{(+10.42)} & 73.57 \textcolor{darkred}{(-2.01)} & 64.38 \textcolor{darkgreen}{(+15.92)} \\
NAC & 39.67 \textcolor{darkgreen}{(+0.05)} & 34.50 \textcolor{darkred}{(-2.82)} & 36.18 \textcolor{darkred}{(-3.44)} & 42.66 \textcolor{darkgreen}{(+5.34)} \\
NMD & 72.78 \textcolor{darkred}{(-6.53)} & 70.68 \textcolor{darkred}{(-3.05)} & 75.08 \textcolor{darkred}{(-4.23)} & 74.22 \textcolor{darkgreen}{(+0.49)} \\
NuSA & 64.02 \textcolor{darkred}{(-11.00)} & 71.26 \textcolor{darkred}{(-3.71)} & 58.45 \textcolor{darkred}{(-16.57)} & 72.52 \textcolor{darkred}{(-2.45)} \\
PCX & 77.34 \textcolor{darkgreen}{(+1.73)} & 68.93 \textcolor{darkgreen}{(+4.18)} & 75.24 \textcolor{darkred}{(-0.37)} & 66.18 \textcolor{darkgreen}{(+1.43)} \\
Residual & 60.22 \textcolor{darkred}{(-5.78)} & 58.69 \textcolor{darkgreen}{(+0.42)} & 69.53 \textcolor{darkgreen}{(+3.53)} & 64.29 \textcolor{darkgreen}{(+6.02)} \\
TAPUUD & 79.66 \textcolor{darkgreen}{(+8.87)} & 67.34 \textcolor{darkgreen}{(+10.33)} & 80.22 \textcolor{darkgreen}{(+9.43)} & 69.30 \textcolor{darkgreen}{(+12.29)} \\
XOOD-M & 84.14 \textcolor{darkgreen}{(+3.38)} & 76.19 \textcolor{darkgreen}{(+12.28)} & 83.92 \textcolor{darkgreen}{(+3.16)} & 76.12 \textcolor{darkgreen}{(+12.21)} \\
\midrule
\multicolumn{5}{l}{\textbf{Confidence-based Methods}} \\
ASH & 53.29 \textcolor{darkred}{(-19.12)} & 43.48 \textcolor{darkred}{(-29.54)} & 57.18 \textcolor{darkred}{(-15.23)} & 42.76 \textcolor{darkred}{(-30.26)} \\
Deep Ensemble & 64.13 \textcolor{darkred}{(-8.93)} & 73.38 \textcolor{darkgreen}{(+0.96)} & 59.91 \textcolor{darkred}{(-13.15)} & 72.19 \textcolor{darkred}{(-0.23)} \\
DICE & 61.82 \textcolor{darkred}{(-7.22)} & 71.59 \textcolor{darkgreen}{(+0.29)} & 58.14 \textcolor{darkred}{(-10.90)} & 71.56 \textcolor{darkgreen}{(+0.26)} \\
GAIA-A & 49.24 \textcolor{darkred}{(-19.69)} & 52.93 \textcolor{darkred}{(-12.62)} & 47.23 \textcolor{darkred}{(-21.70)} & 49.65 \textcolor{darkred}{(-15.90)} \\
GradNorm & 62.81 \textcolor{darkred}{(-12.82)} & 71.27 \textcolor{darkred}{(-1.16)} & 69.09 \textcolor{darkred}{(-6.54)} & 75.61 \textcolor{darkgreen}{(+3.18)} \\
GradOrth & 63.76 \textcolor{darkred}{(-9.04)} & 71.88 \textcolor{darkred}{(-0.86)} & 58.43 \textcolor{darkred}{(-14.37)} & 72.06 \textcolor{darkred}{(-0.68)} \\
MCP & 63.50 \textcolor{darkred}{(-6.33)} & 71.80 \textcolor{darkgreen}{(+3.06)} & 58.42 \textcolor{darkred}{(-11.41)} & 71.65 \textcolor{darkgreen}{(+2.91)} \\
MC-Dropout & 68.34 \textcolor{darkred}{(-2.08)} & 72.81 \textcolor{darkgreen}{(+3.60)} & 58.23 \textcolor{darkred}{(-12.19)} & 71.03 \textcolor{darkgreen}{(+1.82)} \\
ODIN & 63.70 \textcolor{darkred}{(-9.06)} & 72.24 \textcolor{darkred}{(-0.12)} & 58.42 \textcolor{darkred}{(-14.34)} & 71.65 \textcolor{darkred}{(-0.71)} \\
ReAct & 62.84 \textcolor{darkred}{(-1.33)} & 70.46 \textcolor{darkgreen}{(+10.96)} & 55.62 \textcolor{darkred}{(-8.55)} & 67.72 \textcolor{darkgreen}{(+8.22)} \\
SHE & 62.68 \textcolor{darkred}{(-9.52)} & 71.02 \textcolor{darkred}{(-1.34)} & 57.79 \textcolor{darkred}{(-14.41)} & 72.58 \textcolor{darkgreen}{(+0.22)} \\
ViM & 63.36 \textcolor{darkred}{(-11.93)} & 71.43 \textcolor{darkred}{(-3.09)} & 61.28 \textcolor{darkred}{(-14.01)} & 73.67 \textcolor{darkred}{(-0.85)} \\
WeiPer & 65.74 \textcolor{darkred}{(-8.85)} & 66.23 \textcolor{darkred}{(-7.54)} & 72.31 \textcolor{darkred}{(-2.28)} & 73.00 \textcolor{darkred}{(-0.77)} \\

\bottomrule
\end{tabular}
\label{Tbl:cj_resnet}
\end{table*}

\begin{table*}[h]
\centering
\small
\caption{OOD AUROC (\%) on the ISIC ink benchmark using 10 \textbf{ViT-B/32} primary models, trained with light \textbf{colour jitter} augmentations. Brackets show the OOD AUROC difference compared to the primary model's trained without colour jitter augmentations (Table~\ref{Tbl:ISIC_vit}).}
\begin{tabular}{lc >{\columncolor[gray]{0.93}}c }
\toprule
\multirow{2}{*}{\textbf{OOD Method}} & \multicolumn{2}{c}{\textbf{Light Augmentation}}  \\
\cmidrule(lr){2-3}
& Similar & Dissimilar \\
\midrule
\multicolumn{2}{l}{\textbf{Feature-based Methods }}  \\
CoP & 87.82 \textcolor{darkgreen}{(+10.39)} & 72.23 \textcolor{darkgreen}{(+0.15)} \\
CoRP & 80.95 \textcolor{darkgreen}{(+3.44)} & 69.28 \textcolor{darkred}{(-0.77)} \\
FeatureNorm & 82.09 \textcolor{darkgreen}{(+13.72)} & 73.27 \textcolor{darkgreen}{(+20.94)} \\
GRAM & 44.24 \textcolor{darkred}{(-19.30)} & 44.65 \textcolor{darkred}{(-8.03)} \\
KDE (Gaussian) & 85.33 \textcolor{darkgreen}{(+11.44)} & 75.05 \textcolor{darkgreen}{(+10.91)} \\
KNN & 86.76 \textcolor{darkgreen}{(+4.54)} & 79.98 \textcolor{darkgreen}{(+9.65)} \\
LOF & 93.95 \textcolor{darkgreen}{(+2.66)} & 81.13 \textcolor{darkgreen}{(+4.63)} \\
Mahalanobis & 85.06 \textcolor{darkgreen}{(+5.57)} & 79.63 \textcolor{darkgreen}{(+10.24)} \\
MBM & 84.56 \textcolor{darkgreen}{(+3.36)} & 79.21 \textcolor{darkgreen}{(+10.71)} \\
NAN & 56.24 \textcolor{darkgreen}{(+3.17)} & 55.84 \textcolor{darkgreen}{(+3.82)} \\
NAC & 43.19 \textcolor{darkred}{(-2.67)} & 45.62 \textcolor{darkgreen}{(+0.30)} \\
NMD & 80.99 \textcolor{darkgreen}{(+23.11)} & 74.81 \textcolor{darkgreen}{(+17.28)} \\
NuSA & 52.14 \textcolor{darkred}{(-17.83)} & 71.97 \textcolor{darkgreen}{(+3.61)} \\
PCX & 63.40 \textcolor{darkgreen}{(+2.92)} & 59.76 \textcolor{darkgreen}{(+2.97)} \\
Residual & 96.80 \textcolor{darkgreen}{(+16.08)} & 86.46 \textcolor{darkgreen}{(+12.60)} \\
TAPUUD & 90.20 \textcolor{darkgreen}{(+32.50)} & 75.35 \textcolor{darkgreen}{(+20.08)} \\
XOOD-M & 86.33 \textcolor{darkgreen}{(+20.98)} & 68.82 \textcolor{darkgreen}{(+12.11)} \\
\midrule
\multicolumn{2}{l}{\textbf{Confidence-based Methods}} \\
ASH & 51.61 \textcolor{darkred}{(-17.88)} & 64.85 \textcolor{darkred}{(-5.30)} \\
Deep Ensemble & 49.72 \textcolor{darkred}{(-25.76)} & 70.04 \textcolor{darkred}{(-2.54)} \\
DICE & 76.39 \textcolor{darkgreen}{(+8.47)} & 66.25 \textcolor{darkred}{(-1.46)} \\
GAIA-A & 49.46 \textcolor{darkgreen}{(+14.68)} & 30.10 \textcolor{darkred}{(-1.09)} \\
GradNorm & 57.27 \textcolor{darkred}{(-13.92)} & 69.96 \textcolor{darkred}{(-0.74)} \\
GradOrth & 77.28 \textcolor{darkgreen}{(+9.00)} & 73.72 \textcolor{darkgreen}{(+3.52)} \\
MCP & 48.82 \textcolor{darkred}{(-20.36)} & 69.40 \textcolor{darkgreen}{(+0.32)} \\
MC-Dropout & 58.28 \textcolor{darkred}{(-14.04)} & 58.60 \textcolor{darkred}{(-13.54)} \\
ODIN & 48.82 \textcolor{darkred}{(-22.51)} & 70.43 \textcolor{darkgreen}{(+0.08)} \\
ReAct & 38.80 \textcolor{darkred}{(-29.09)} & 53.37 \textcolor{darkred}{(-12.79)} \\
SHE & 57.67 \textcolor{darkred}{(-12.31)} & 75.48 \textcolor{darkgreen}{(+5.85)} \\
ViM & 39.83 \textcolor{darkred}{(-30.82)} & 57.24 \textcolor{darkred}{(-12.90)} \\
WeiPer & 73.18 \textcolor{darkgreen}{(+3.20)} & 70.95 \textcolor{darkgreen}{(+4.64)} \\

\bottomrule
\end{tabular}

\label{Tbl:cj_vit}
\end{table*}

\begin{table}[h]
\centering
\caption{Average inference latency (in milliseconds) for a single image on the ISIC Ink benchmark on a ResNet18 primary model. Results are shown for three feature-based OOD detection methods with and without the proposed subspace projection, showing small increases in inference latency. The forward pass through the model is included in this value. The inference latency of the external method baseline (DDPM-MSE) is included to highlight that some OOD detection methods have significantly higher computational cost. }
\begin{tabular}{|l|l|}
\hline
\textbf{Method} & \textbf{Inference Latency (ms)} \\
\hline
Mahalanobis & 12.15  \\
+ proj. & 14.23 \\
\hline
Featurenorm & 8.63  \\
+ proj. & 10.32 \\
\hline
NaN & 11.04  \\
+ proj. & 13.09 \\
\hline
DDPM-MSE & 4767.90 \\
\bottomrule
\end{tabular}

\label{tab:inf_lat}
\end{table}

\begin{table}[h]
\centering
\caption{Average OOD detection AUROC (\%) for the Mahalanobis method as a function of the nuisance subspace dimensionality $k$. Removing a small number of nuisance directions (up to $k=5$) improves performance for both similar and dissimilar artefacts, whereas removing larger subspaces ($k \geq 10$) removes task-relevant structure and substantially degrades AUROC.
}
\begin{tabular}{|l| c| c | c | c |}
\hline
\textbf{OOD Artefact} & \textbf{k=2} & \textbf{k=5}  & \textbf{k=10} & \textbf{k=20} \\

\hline

Similar & 77.2 & 77.5 & 53.7 & 48.2 \\
\rowcolor{gray!20}
Dissimilar & 67.0 & 75.8 & 59.4 & 47.0 \\
\bottomrule
\end{tabular}

\label{tab:k_hyper}
\end{table}

\begin{table}[h]
\centering
\caption{OOD performance on ISIC ink and colour chart benchmarks with and without nuisance subspace projection (equation 3), averaged over 25 ResNet18 models with random seeds. Using projected features can reduce the Invisible Gorilla Effect across several feature based OOD detection methods, although not for all methods (e.g. XOOD-M).  
}
\begin{tabular}{l c >{\columncolor[gray]{0.93}}c c >{\columncolor[gray]{0.93}}c}
    \toprule
   \multirow{2}{*}{\textbf{Method}} & \multicolumn{2}{c}{\textbf{Ink Artefacts}} & \multicolumn{2}{c}{\textbf{Colour Charts}} \\ \cmidrule(lr){2-3} \cmidrule(lr){4-5}
  & Sim. & Diss. & Sim. & Diss.\\
    \midrule
    Mahalanobis & 77.0 & 63.6 & 96.7 & 95.4  \\
    +Proj.   & 77.5 & 75.8 & 95.8 & 97.7 \\
    \midrule
    MBM &  77.4 & 63.7 & 97.0 & 95.5  \\
    +Proj.   & 77.6 & 75.3 & 96.1 & 97.2 \\
    \midrule
    FeatureNorm & 75.1 & 52.9 & 62.4 & 58.1 \\
    +Proj.   & 75.3 & 74.5 & 75.9 & 76.3 \\
    \midrule
    NAN & 75.6 & 48.5 & 72.5 & 68.1 \\
    +Proj.   & 75.3 & 76.8 & 77.9 & 77.4 \\
    \midrule
    XOOD-M & 80.8 & 63.9 & 88.5 & 84.8 \\
    +Proj.   & 68.5 & 61.0 & 61.2 & 59.8 \\
    \bottomrule
    \end{tabular}

\label{tab:proj_res}
\end{table}

\end{document}